\documentclass[11pt,letterpaper]{article}

\usepackage{graphicx}  
\usepackage{amsmath}   
\usepackage{amsthm}
\usepackage{amssymb}   
\usepackage{geometry}  
\usepackage{authblk}   
\usepackage{natbib}    
\usepackage{enumitem}  
\usepackage{amsfonts}
\usepackage{bbm}
\usepackage{mathtools}
\usepackage{makecell}
\usepackage{tikz}
\usepackage{bm}
\usepackage{array}
\usepackage{tabularx}
\usepackage{xcolor}
\usepackage{microtype}

\usepackage[suppress]{color-edits} 
\addauthor[sz]{sz}{blue}
\addauthor[claude]{claude}{red}

\usepackage[vlined, ruled, linesnumbered]{algorithm2e}
\usepackage{etoolbox}
\AtBeginEnvironment{algorithm}{\normalsize}

\usepackage{booktabs}
\usepackage{lineno}

\usepackage{hyperref}  

\newtheorem{lemma}{Lemma}

\newtheorem{theorem}{Theorem}

\newtheorem{proposition}{Proposition}

\newtheorem{corollary}{Corollary}

\usepackage{cleveref}

\crefname{theorem}{Theorem}{Theorems}
\Crefname{theorem}{Theorem}{Theorems}
\crefname{lemma}{Lemma}{Lemmas}
\Crefname{lemma}{Lemma}{Lemmas}
\crefname{proposition}{Proposition}{Propositions}
\Crefname{proposition}{Proposition}{Propositions}
\crefname{corollary}{Corollary}{Corollaries}
\Crefname{corollary}{Corollary}{Corollaries}
\crefname{remark}{Remark}{Remarks}
\Crefname{remark}{Remark}{Remarks}
\crefname{definition}{Definition}{Definitions}
\Crefname{definition}{Definition}{Definitions}

\newcommand{\Reg}{\mathrm{Reg}}

\newcommand{\TV}{\mathrm{TV}}

\newcommand{\E}{\mathbb{E}}
\newcommand{\Prb}{\mathbb{P}}

\newcommand{\1}{\mathbf{1}}

\newcommand{\cF}{\mathcal{F}}
\newcommand{\cA}{\mathcal{A}}

\newcommand{\KL}{D_{\mathrm{KL}}}
\newcommand{\TVdist}{d_{\mathrm{TV}}}

\newcommand{\eps}{\varepsilon}

\newcommand{\order}{\preceq}

\newcommand{\PiMass}{\Pi}

\newcommand{\wtO}{\widetilde O}
\newcommand{\wtTheta}{\widetilde\Theta}

\crefname{algocf}{Algorithm}{Algorithms}
\Crefname{algocf}{Algorithm}{Algorithms}

\SetKwFunction{Pair}{Pair}
\SetKwFunction{Est}{Estimate}
\SetKwProg{Fn}{Procedure}{:}{}

\title{Online Learning with Ordered Experts (via two-point feedback)}
\author[1]{Shiliang Zuo}
\affil[1]{Meta, szuo.rs@gmail.com}
\date{}

\begin{document}

\maketitle  

\begin{abstract}
\normalsize
We study adversarial online learning with expert recommendations in a totally ordered action space, where the experts' recommendations may change and be reordered arbitrarily. Our focus is two-point feedback: on each round the learner plays two current recommendations, observes both, and pays their average. We study two problems. In the separated monotone problem, the loss is the sum of a hidden nondecreasing component and a hidden nonincreasing component, and a query reveals the two components separately. In the BV problem, the losses are bounded and have bounded variation (BV) along the order, and a query reveals only the loss.

In the separated monotone problem, two paid observations achieve
\[
O\bigl(\log(|\cF|T)\sqrt{T\log|\cF|}\bigr)
\]
regret against the best of $|\cF|$ experts, within a logarithmic factor of the classical full-information bound. For the BV problem, we obtain $\wtO(\sqrt T+|\cF|)$ regret.
We give an application in dynamic pricing with expert advice, showing that two-option menus matches the full information rate up to $\log$ factors. 

The main technical ingredient is a marginal-preserving endpoint flow estimator. Two observations reveal a loss increment along the order; random paths supply telescoping identities that turn this local information into unbiased loss estimates for all current recommendations. A reversible pair sampler preserves the learner's intended marginal at both played actions. The estimator's second-moment and stability bounds support online updates for both feedback models.

\end{abstract}
\clearpage

\section{Introduction}
\label{sec:introduction}

In online learning, a learner makes a decision on every round, an adversary chooses the loss of every possible decision, and the learner aims to keep its cumulative loss close to that of the best strategy in a comparison class. We study the expert-advice version of this problem. On every round, each expert in a finite class $\cF$ recommends an action, the learner chooses an action, and its regret is measured against the best expert in hindsight. In the classical experts and bandit problems, the actions form an unstructured set, and nothing relates the losses of different actions. In many applications, however, the actions carry a natural total order.

For example, in pricing, a seller chooses a price. In inventory management, a retailer chooses a stock level. In classification, a decision maker chooses a threshold and flags every score above it. In each case the experts' recommendations are points on a line, and they may change and be reordered arbitrarily across rounds.

The order brings additional structure that a learner may exploit. Many problems involve a natural tradeoff between two monotone losses: moving the action up increases one cost and decreases the other. The loss then decomposes as
\[
\ell_t=\ell_t^{\uparrow}+\ell_t^{\downarrow},
\]
with a nondecreasing component $\ell_t^{\uparrow}$ and a nonincreasing component $\ell_t^{\downarrow}$.

We formulate the problem of online learning with ordered experts and focus on two-point feedback. The learner simultaneously plays two current recommendations, observes the feedback at both, and pays the average of their losses; there are no free probes. We study two problems. In the \emph{separated monotone problem}, each loss is a monotone sum as above, and a query at an action reveals $\ell_t^{\uparrow}$ and $\ell_t^{\downarrow}$ there separately. In the \emph{BV problem}, each loss has total variation at most one along the order, and a query reveals only the loss. On a finite chain, the Jordan decomposition expresses every loss of bounded variation (BV) as such a monotone sum, up to an action-independent offset and a constant rescaling, so the two problems involve essentially the same losses and differ only in what a query reveals. To calibrate the results, we compare with one-point feedback, where the learner plays and observes a single recommendation, and with full information.

At a high level, two observations on the same round reveal an increment of the loss along the order, which no single observation can, and this is what separates two-point from one-point feedback. In the separated monotone problem, two observations match full information up to a logarithmic factor, whereas one observation costs a polynomial factor. In the BV problem, two observations replace the multiplicative $\sqrt{|\cF|}$ cost of bandit feedback by an additive $|\cF|$.

An important application is dynamic pricing, with or without expert advice. Menus with two purchase options and an outside option for one risk-neutral buyer yield revenue-regret guarantees against the best fixed posted price and against the best price-recommending expert.

\subsection{Results}

\begin{table}[t]
\centering
\normalsize
\renewcommand{\arraystretch}{1.18}
\begin{tabularx}{\textwidth}{@{}>{\raggedright\arraybackslash}p{0.18\textwidth}>{\raggedright\arraybackslash}p{0.32\textwidth}>{\raggedright\arraybackslash}p{0.25\textwidth}>{\raggedright\arraybackslash}X@{}}
\toprule
Problem & \textbf{two paid observations} & one observation & full information \\
\midrule
Separated monotone problem
& $O(\log(|\cF|T)\sqrt{T\log|\cF|})$ (\Cref{thm:separated-upper}); $\Omega(\sqrt T\log\min\{|\cF|,\sqrt T\})$ (\Cref{thm:dyadic-lower})
& $\Omega(\min\{\sqrt{|\cF|T},T^{2/3}\})$ (\Cref{thm:separated-one-lower})
& $\Theta(\sqrt{T\log|\cF|})$ (\Cref{thm:full-info}) \\
\addlinespace
BV problem
& $\wtTheta(\sqrt T+|\cF|)$ \par(\Cref{cor:BV-two-point-rate})
& $\Theta(\sqrt{|\cF|T})$ \par(\Cref{thm:BV-onepoint})
& $\Theta(\sqrt{T\log|\cF|})$ (\Cref{thm:full-info}) \\
\bottomrule
\end{tabularx}
\caption{Regret against the best of $|\cF|$ experts. The BV two-point and one-point $\Theta$ rates are stated for $|\cF|\le T$; the full-information $\Theta$ rate is stated for $\log_2|\cF|\le T$. All lower bounds hold even when every expert recommends one fixed action on every round.}
\label{tab:results}
\end{table}

\paragraph{The separated monotone problem.}
Against $|\cF|$ experts with arbitrary current recommendations, two paid observations achieve
\[
\E\Reg_T(\cF)
=
O\bigl(\log(|\cF|T)\sqrt{T\log|\cF|}\bigr).
\]
The key ingredient is the endpoint flow estimator described below: from the two observed values it builds unbiased and nonnegative estimates of both monotone parts at every active recommendation, so exponential weights can be run on them. Conversely, no two-point learner can beat $\Omega(\sqrt T\log\min\{|\cF|,\sqrt T\})$, even when the experts never move; this lower bound holds in the BV problem as well.

\paragraph{The BV problem.}
When a query reveals only the total loss and its variation is at most one, two paid observations achieve $\wtO(\sqrt T+|\cF|)$ regret. Signed endpoint-flow estimates are controlled by an entropy--log-barrier update using their coordinatewise bound. Matching lower bounds up to logarithmic factors show that the two-point rate is $\wtTheta(\sqrt T+|\cF|)$ when $|\cF|\le T$. 

\paragraph{Dynamic pricing with expert advice. }
We apply our framework to adversarial dynamic pricing through an incentive-compatible menu mechanism. On each round, the two prices sampled by our algorithm are converted into a menu with two purchase options and an outside option. The buyer's choice reveals the acceptance decisions at both prices, while the seller's expected revenue equals the average revenue of the two posted prices. This implementation yields $O(\sqrt{T}\log^{3/2}T)$ expected revenue regret against the best fixed posted price. More generally, given a finite class $\mathcal{F}$ of price-recommending experts, we achieve $O(\log(|\mathcal{F}|T)\sqrt{T\log|\mathcal{F}|})$ regret against the best expert, even when expert recommendations change and reorder arbitrarily across rounds. This matches the full information up to $\log$ factors. For comparison, the one-point bandit (posted-price auction) feedback gives regret $\Theta(T^{2/3})$ against the best fixed price setting. 

\paragraph{Comparison with one point and full information.}
As summarized in \Cref{tab:results}, one observation gives regret $\Omega(\min\{\sqrt{|\cF|T},T^{2/3}\})$ in the separated monotone problem and $\Theta(\sqrt{|\cF|T})$ in the BV problem for $|\cF|\le T$. Full information gives $\Theta(\sqrt{T\log|\cF|})$ in both problems for $\log_2|\cF|\le T$, even with fixed recommendations. For $|\cF|\le T^{1/3}$, two-point feedback removes the one-point rate's multiplicative $\sqrt{|\cF|}$ dependence up to logarithms; full information also removes the additive $|\cF|$ term in our BV upper bound. 

\subsection{The endpoint flow estimator}

All of our two-point upper bounds rest on a single estimator. Standard one-point importance weighting preserves unbiasedness by dividing the observed loss by its sampling probability, but its distribution-weighted second moment can grow linearly with the number of distinct recommendations.

Under two-point feedback, our key technical tool is the \emph{endpoint flow estimator} (\Cref{thm:endpoint-flow}). For a target vector $g$ with values in $[0,1]$ and total variation at most one, let $\pi$ be the distribution over distinct current recommendations after the $1/T$ smoothing used by our algorithms. Three properties are central:
\begin{enumerate}[nosep,leftmargin=2em]
\item Both played actions have marginal $\pi$.
\item It is unbiased at every coordinate: $\E\widehat g(j)=g(j)$.
\item Its distribution-weighted second moment satisfies $\E\sum_j\pi(j)\widehat g(j)^2=O(\log^2(|\cF|T))$.
\end{enumerate}
Marginal preservation aligns the expected paid loss with the intended mixture, while unbiasedness and the second-moment bound control estimation error. \Cref{sec:endpoint-flow} gives the construction and full analysis.

\subsection{Related work}

\paragraph{Expert advice and ordered actions.} Exponential weights gives the full-information benchmark \citep*{littlestone1994weighted,freund1997decision}; see \citet*{cesa2006prediction}. One-point expert advice uses the bandit framework \citep*{auer2002nonstochastic,mcmahan2009tighter}, with ordinary bandit rates providing the comparison for our BV result \citep*{audibert2009,bubeck2012,lattimore2020bandit}. Metric bandits \citep*{kleinberg2008multi}, unimodal bandits \citep*{yu2011unimodal,combes2014unimodal}, posted pricing \citep*{kleinberg2003value}, and censored-demand inventory control \citep*{huh2009nonparametric} exploit structure over a fixed action set. Here experts may change and reorder their recommendations arbitrarily.

\paragraph{Multi-point feedback.} Single-point convex-bandit estimates \citep*{flaxman2005online} can be improved using two-point queries \citep*{agarwal2010optimal,duchi2015optimal,shamir2017optimal}. Feedback graphs \citep*{mannor2011bandits,alon2015} and paid side observations \citep*{seldin2014prediction} provide information beyond the played action; dueling bandits observe comparisons \citep*{yue2012k}. Our two played actions supply numerical interval increments for reconstructing the ordered loss vector, with their intended marginals preserved.

\paragraph{Regularization and reversible sampling.} Log-barrier and hybrid entropy--log-barrier updates stabilize importance-weighted estimates \citep*{foster2016learning,agarwal2017corralling,wei2018more,bubeck2018sparsity}; endpoint flow supplies the required coordinatewise control. The pair sampler uses a reversible kernel with prescribed stationary distribution \citep*{hastings1970monte}, with conductances chosen to control reconstruction variance.

\paragraph{Dynamic pricing and menus.}
Classical dynamic pricing learns a posted price from binary purchase feedback, with $\widetilde{\Theta}(T^{2/3})$ regret against the best fixed price under adversarial valuations \citep{kleinberg2003value}. Contextual pricing extends this setting by incorporating observable features, typically under structural assumptions on valuations or demand \citep{liu2021optimal,xu2024pricing}. Menus have also been studied as revenue-maximizing mechanisms \citep{balcan2024menus} and as a means of improving regret against strategic buyers \citep{zuo2026menus}. In contrast, we consider fresh, myopic buyers with obliviously adversarial valuations and use a menu with two purchase options to obtain two price observations from each buyer without sacrificing expected revenue relative to their average. This yields $\widetilde O(\sqrt{T\log|\mathcal F|})$ regret, up to logarithmic factors, against the best expert from a finite class of potentially time-varying pricing policies, with the classical fixed-price benchmark as a special case.

\subsection{Organization}
\Cref{sec:model} defines the ordered-expert model, the two-point protocol, and the two problems. \Cref{sec:endpoint-flow} develops the marginal-preserving endpoint flow estimator. \Cref{sec:separated} presents and analyzes a two-point algorithm for the separated monotone problem and proves a lower bound. \Cref{sec:bv} presents and analyzes a two-point algorithm for the BV problem and proves lower bounds that match it up to logarithmic factors. \Cref{sec:menu-application} contains a mechanism-design native application in dynamic pricing. \Cref{sec:one-point-full-info} contains the one-point and full-information comparisons. Proofs of the general results are deferred to the appendices.

\section{Model and preliminaries}
\label{sec:model}

\subsection{Ordered experts}

Fix a horizon $T\ge2$. Let $(\cA,\order)$ be a totally ordered action space with a least element $\bot$ and a greatest element $\top$. For example, $\cA=[0,1]$ with the usual order, so that $\bot=0$ and $\top=1$, where an action may be a normalized price, a threshold, or a stock level. Let $\cF$ be a finite class of $|\cF|\ge2$ experts. On round $t$, before the learner acts, every expert $f\in\cF$ recommends an action $a_{t,f}\in\cA$. The recommendations may change arbitrarily across rounds. The oblivious adversary fixes all recommendations and all loss functions $\ell_t:\cA\to[0,1]$ before the interaction begins. Let
\[
X_t=\{a_{t,f}:f\in\cF\}=\{x_{t,1}\prec\cdots\prec x_{t,m_t}\},
\qquad m_t\le|\cF|,
\]
be the distinct active actions at round $t$. We call $j$ the \emph{rank} of $x_{t,j}$, write $j_t(f)$ for the rank of the recommendation $a_{t,f}$, and identify each active action with its rank when convenient; in particular, a distribution over $X_t$ is a distribution over the ranks $\{1,\ldots,m_t\}$.

We study two problems, which differ in what a query reveals. Before that, we define the notion of total variation. For a function $g$ on a finite chain $x_1\prec\cdots\prec x_m$,
\[
\TV(g)=\sum_{j=2}^m|g(x_j)-g(x_{j-1})|
\]
is its total variation. Unless stated otherwise, $\TV(\ell_t)$ denotes the total variation of $\ell_t$ restricted to the active chain $X_t$.

\paragraph{The separated monotone problem: monotone-sum losses, separated observations.} On each round the adversary chooses a nondecreasing component $\ell_t^{\uparrow}:\cA\to[0,1/2]$ with $\ell_t^{\uparrow}(\bot)=0$ and a nonincreasing component $\ell_t^{\downarrow}:\cA\to[0,1/2]$ with $\ell_t^{\downarrow}(\top)=0$, and the loss is their sum,
\[
\ell_t(a)=\ell_t^{\uparrow}(a)+\ell_t^{\downarrow}(a).
\]
Querying an action $a$ reveals $\ell_t^{\uparrow}(a)$ and $\ell_t^{\downarrow}(a)$ separately.

\paragraph{The BV problem: BV losses, aggregate observations.} On every round the loss $\ell_t:\cA\to[0,1]$ has bounded variation, $\TV(\ell_t)\le1$, and querying an action $a$ reveals only $\ell_t(a)$.

\paragraph{Relation between the two problems.} The two problems involve essentially the same losses and differ in what a query reveals. Every loss of the separated monotone problem is a BV loss, since $\TV(\ell_t)\le\TV(\ell_t^{\uparrow})+\TV(\ell_t^{\downarrow})\le1/2+1/2$. Conversely, by the Jordan decomposition, every loss with $\TV(\ell_t)\le1$ can be written as
\[
\ell_t(x)=c_t+\ell_t^{\uparrow}(x)+\ell_t^{\downarrow}(x),
\qquad x\in X_t,
\]
where the offset $c_t$ is a constant that does not depend on $x$. On the active chain, take
\[
\ell_t^{\uparrow}(x_{t,j})=\sum_{r=2}^{j}\bigl(\ell_t(x_{t,r})-\ell_t(x_{t,r-1})\bigr)^+,
\qquad
\ell_t^{\downarrow}(x_{t,j})=\sum_{r=j+1}^{m_t}\bigl(\ell_t(x_{t,r-1})-\ell_t(x_{t,r})\bigr)^+,
\]
where $z^+=\max\{z,0\}$, and set $c_t=\ell_t(x_{t,1})-\ell_t^{\downarrow}(x_{t,1})$. Extend the components to $\cA$ as step functions, keeping the upward component zero below the first active action and the downward component zero above the last. Thus $\ell_t^{\uparrow}$ is nondecreasing with $\ell_t^{\uparrow}(\bot)=0$, $\ell_t^{\downarrow}$ is nonincreasing with $\ell_t^{\downarrow}(\top)=0$, and $\ell_t^{\uparrow}(\top)+\ell_t^{\downarrow}(\bot)=\TV(\ell_t)\le1$. The offset $c_t$ shifts every expert's loss equally and does not affect regret. Hence, after removing the offset and dividing both components by two, the active loss vector belongs to the separated monotone problem. This is a relation between loss classes; aggregate feedback does not reveal the decomposition or its offset.

\subsection{Two-point feedback and regret}

Our main protocol gives the learner two paid observations per round. The learner simultaneously chooses $A_t,B_t\in X_t$ and incurs
\[
\frac12\bigl(\ell_t(A_t)+\ell_t(B_t)\bigr).
\]
For the separated monotone problem, the learner then observes the four values $\ell_t^{\uparrow}(A_t)$, $\ell_t^{\downarrow}(A_t)$, $\ell_t^{\uparrow}(B_t)$, and $\ell_t^{\downarrow}(B_t)$. For the BV problem, it observes only the two values $\ell_t(A_t)$ and $\ell_t(B_t)$. The regret to the best expert is
\[
\Reg_T(\cF)
=
\sum_{t=1}^T\frac{\ell_t(A_t)+\ell_t(B_t)}2
-
\min_{f\in\cF}\sum_{t=1}^T\ell_t(a_{t,f}).
\]
We bound the expected regret $\E\Reg_T(\cF)$, where the expectation is over learner randomness. We write $\E_t$ for expectation conditional on the history before the learner draws its pair on round $t$. \Cref{sec:one-point-full-info} compares two-point feedback with one-point feedback, where the learner plays and observes a single action, and with full information.

\section{The endpoint flow estimator}
\label{sec:endpoint-flow}

This section develops the estimator behind our main results. Throughout this section, we fix one round and suppress the time index. The active chain is $x_1\prec\cdots\prec x_m$, and $g$ is a hidden vector on it with entries $g(x_1),\ldots,g(x_m)$. For the separated monotone problem, $g$ is one of the two monotone components, $\ell^{\uparrow}$ or $\ell^{\downarrow}$. For the BV problem, $g$ is the loss vector $\ell$. Finally, $\pi$ is the distribution over ranks $\{1,\ldots,m\}$ with which the learner intends to play.

\subsection{Main result and overview}

\paragraph{Main result.}
We assume a floor on the intended distribution,
\[
\pi(j)\ge \frac{\xi}{m},
\qquad j=1,\ldots,m,
\]
where $0<\xi\le1$, and write
\[
\Lambda:=1+\log\frac{m}{\xi}.
\]
Any distribution $\pi^\circ$ on the active ranks can be brought into this form by mixing it with $\xi$ times the uniform distribution on active actions, at a cost of at most $\xi$ per round in expected loss. Our algorithms do this, and the final choice will be $\xi=1/T$, so that $\Lambda\le1+\log(|\cF|T)$ and the mixing costs at most one in total loss over $T$ rounds. The anchor $x_0$ is an auxiliary node below the active chain with assigned value $g(x_0)=0$, even if $x_1=\bot$; it is never played and serves only as the starting point of the reconstruction.

\begin{theorem}[Endpoint flow estimator]
\label{thm:endpoint-flow}
Let $\pi$ satisfy the floor $\pi(j)\ge\xi/m$. There exist a random pair of ranks $(A,B)$, whose law depends only on $\pi$, and an estimator $\widehat g$, which depends on $g$ only through the observed values $g(x_A)$ and $g(x_B)$, such that for every hidden vector $g$ with values in $[0,1]$ and $\TV(g)\le1$:
\begin{enumerate}
\item \textbf{marginals:} $A\sim\pi$ and $B\sim\pi$;
\item \textbf{unbiasedness:} $\E[\widehat g(j)]=g(x_j)$ for every target $j$;
\item \textbf{second moment:}
\[
\E\langle\pi,\widehat g^2\rangle
=
O(\Lambda^2);
\]
\item \textbf{self-bound:} for every realization,
\[
\max_j \pi(j)|\widehat g(j)|
\le
3\Lambda;
\]
\item \textbf{monotone sign:} if $g$ is nondecreasing, then $\widehat g(j)\ge0$ for every $j$ and every realization.
\end{enumerate}
\end{theorem}

\paragraph{High-level overview.}
Fix a target rank $j$. Imagine a particle that starts at the anchor $x_0$ and moves up the chain one jump at a time until it exits at $x_j$; from $x_a$ it jumps to a rank $b\in\{a+1,\ldots,j\}$ with a probability, specified later, that depends only on $\pi$. Let $\phi_j(a,b)$ be the probability that the particle jumps directly from $x_a$ to $x_b$. Along every trajectory the increments $g(x_b)-g(x_a)$ of the jumps telescope to $g(x_j)-g(x_0)=g(x_j)$, so averaging over trajectories gives the decomposition
\[
g(x_j)=\sum_{0\le a<b\le j}\phi_j(a,b)\bigl(g(x_b)-g(x_a)\bigr).
\]
Thus $g(x_j)$ is a fixed linear combination of increments, with coefficients determined by $\pi$ alone.

Now suppose the queried pair is sampled from some probability law. The value $g(x_A)$ alone reveals the anchor increment $g(x_A)-g(x_0)$ on the edge $(0,A)$, and the two values together reveal the increment on the edge between $A$ and $B$. Write $q(a,b)$ for the probability that the increment on the edge $(a,b)$ is revealed. If $q(a,b)>0$ whenever $\phi_j(a,b)>0$, then weighting each revealed increment by $\phi_j(a,b)/q(a,b)$ and summing gives an unbiased estimate of $g(x_j)$, simultaneously for every target $j$, from only two observations.

Two design questions remain. First, the pair cannot follow an arbitrary law: both queried actions are played, so both must have marginal $\pi$. We draw $A\sim\pi$ and then $B$ by one step of a reversible Markov kernel with stationary distribution $\pi$, so that detailed balance gives $B\sim\pi$ as well. Second, the importance weights $\phi_j(a,b)/q(a,b)$ must not inflate the second moment. We therefore match the kernel to the particle: pairs of likely actions separated by little $\pi$-mass, which carry most of the flow, are sampled most often. The next subsection states this importance-weighting step in general, and the one after it gives the specific paths and kernel.

\subsection{Telescoping flows and importance weighting}
\label{sec:endpoint-flow-general}

This subsection builds an unbiased estimate of every value $g(x_j)$ from the two observed values $g(x_A)$ and $g(x_B)$. It works for any law of the queried pair; the specific law is chosen in the next subsection.

\paragraph{Edges.} Index the anchor by rank $0$, so that $g(x_0)=0$. An \emph{edge} is a pair of ranks $e=(a,b)$ with $0\le a<b\le m$. It is an \emph{anchor edge} if $a=0$ and an \emph{interval edge} if $a\ge1$. Its \emph{increment} is
\[
H(e)=g(x_b)-g(x_a).
\]

\paragraph{Flows.} For a target rank $j$, let a random increasing path run from the anchor to $j$, jumping from each rank $a<j$ to some rank $b\in\{a+1,\ldots,j\}$, and let $\phi_j(e)$ be the probability that it traverses the edge $e$. The law of the path depends only on $\pi$ and is specified in the next subsection. Along every realization the increments telescope to $g(x_j)-g(x_0)=g(x_j)$, so taking expectations gives
\[
g(x_j)=\sum_e\phi_j(e)H(e).
\]

\paragraph{Revealed edges.} Write $A\wedge B=\min\{A,B\}$ and $A\vee B=\max\{A,B\}$. Since $g(x_0)=0$, the value $g(x_A)$ alone reveals the increment of the anchor edge $E_1=(0,A)$. If $A\ne B$, the two values together also reveal the increment of the interval edge $E_2=(A\wedge B,A\vee B)$; if $A=B$, no interval edge is revealed. The \emph{revelation probability} $q(e)$ of an edge is the probability that its increment is revealed:
\[
q(0,a)=\Prb(E_1=(0,a)),
\qquad
q(a,b)=\Prb(E_2=(a,b))\quad(1\le a<b).
\]

\begin{lemma}[Unbiased estimator]
\label{lem:unbiased-estimator}
Suppose that $q(e)>0$ whenever $\phi_j(e)>0$, and define
\[
\widehat g(j)
=
\sum_{a}\1\{E_1=(0,a)\}\frac{\phi_j(0,a)}{q(0,a)}H(0,a)
+
\sum_{1\le a<b}\1\{E_2=(a,b)\}\frac{\phi_j(a,b)}{q(a,b)}H(a,b),
\]
where the sums run over edges with $q>0$. Then $\E\widehat g(j)=g(x_j)$ for every $j$. If $g$ is nonnegative and nondecreasing, then $\widehat g(j)\ge0$ for every $j$ and every realization.
\end{lemma}

Each revealed increment is weighted by its flow divided by its revelation probability, so in expectation every edge contributes $\phi_j(e)H(e)$ and the identity above is recovered; the proof is in \Cref{app:endpoint-flow}. The first sum handles the anchor edges and the second the interval edges, so $E_1$ and $E_2$ need not lie on a common path. Two choices remain: the jump probabilities of the path, which determine the flows $\phi_j$, and the law of the queried pair, which determines the revelation probabilities $q$. Both queried actions are played, so both must have marginal $\pi$, and the weights $\phi_j/q$ must keep the second moment of $\widehat g$ small.

\subsection{The construction}
\label{sec:endpoint-flow-construction}

This subsection makes these two choices and bounds the second moment of the resulting estimator. Throughout, write $\PiMass[a,b]=\sum_{r=a}^b\pi(r)$ and $\PiMass(a,b]=\sum_{r=a+1}^b\pi(r)$ for the $\pi$-mass of an interval of ranks.

\paragraph{Random paths.}
For a target rank $j$, the particle of the overview starts at the anchor $0$, and from a node $a<j$ it jumps to $b\in\{a+1,\ldots,j\}$ with probability proportional to $\pi(b)$:
\[
\Prb(b\mid a,j)=\frac{\pi(b)}{\PiMass(a,j]}.
\]
The particle thus steps preferentially onto actions that the learner is likely to play. Its edge-use probabilities, computed in \Cref{app:endpoint-flow}, are
\[
\phi_j(0,a)=\frac{\pi(a)}{\PiMass[1,j]}
\quad(a\le j),
\qquad
\phi_j(a,b)=
\frac{\pi(a)\pi(b)}{\PiMass[a,j]\PiMass(a,j]}
\quad(1\le a<b\le j).
\]
Set $\phi_j(a,b)=0$ for all edges outside $0\le a<b\le j$.

\paragraph{Reversible pair sampler.}
Now we define how the pair of points is actually sampled. For $1\le a<b\le m$, define the symmetric conductance
\[
\Gamma(a,b)
=
\frac{1}{4\Lambda}
\frac{\pi(a)\pi(b)}{\PiMass[a,b]},
\]
set $\Gamma(b,a)=\Gamma(a,b)$, and define the kernel
\[
\kappa(b,a)=\frac{\Gamma(a,b)}{\pi(b)}\quad(a\ne b),
\qquad
\kappa(b,b)=1-\sum_{a\ne b}\kappa(b,a).
\]
The conductance is largest for pairs of likely actions separated by little $\pi$-mass. These are the edges on which the random paths place the most flow, so the kernel observes each edge roughly in proportion to its use; the factor $1/(4\Lambda)$ keeps every row of $\kappa$ a probability distribution.

\begin{lemma}[Kernel validity]
\label{lem:kernel-validity}
The kernel $\kappa$ is valid, in particular $\kappa(b,b)\ge1/2$, and satisfies detailed balance
\[
\pi(b)\kappa(b,a)=\pi(a)\kappa(a,b)=\Gamma(a,b).
\]
Hence $\pi$ is stationary.
\end{lemma}

Draw $A\sim\pi$ and then $B\sim\kappa(A,\cdot)$. By stationarity, $B\sim\pi$ as well, which is the marginal claim of the theorem; in particular, the expected paid average is $\langle\pi,g\rangle$. For this pair, the revelation probabilities are
\[
q(0,a)=\pi(a),
\qquad
q(a,b)=\pi(a)\kappa(a,b)+\pi(b)\kappa(b,a)=2\Gamma(a,b)\quad(1\le a<b).
\]

\paragraph{Estimator.}
The endpoint flow estimate is the estimator of \Cref{lem:unbiased-estimator} for this pair and the flows above. Since $H(0,A)=g(x_A)$ and $q(0,A)=\pi(A)$, it reads
\[
\widehat g(j)
=
\frac{\phi_j(0,A)}{\pi(A)}\,g(x_A)
+
\1\{A\ne B\}
\frac{\phi_j(A\wedge B,A\vee B)}{q(A\wedge B,A\vee B)}
\bigl(g(x_{A\vee B})-g(x_{A\wedge B})\bigr).
\]
When $A=B$, the interval term is omitted; no flow or revelation probability is evaluated on a diagonal edge. The estimate uses only the two observed values and the known distribution $\pi$. By that lemma, $\widehat g$ is unbiased. The flows are nonnegative, and a nondecreasing $g$ with values in $[0,1]$ has nonnegative increments, including the anchor increments $g(x_a)-g(x_0)=g(x_a)$, so $\widehat g\ge0$ in that case. \Cref{alg:endpoint-flow} summarizes the sampler and the estimator; our learning algorithms call its two procedures.

\begin{algorithm}[t]
\caption{Endpoint flow}
\label{alg:endpoint-flow}
\Fn{\Pair{$\pi$}}{
Draw $A\sim\pi$ and then $B\sim\kappa(A,\cdot)$, where $\kappa(b,a)=\Gamma(a,b)/\pi(b)$ for $a\ne b$ and $\kappa(b,b)=1-\sum_{a\ne b}\kappa(b,a)$\;
\Return{$(A,B)$}\;
}
\Fn{\Est{$\pi,A,B,g(x_A),g(x_B)$}}{
\For{$j=1,\ldots,m$}{
$\widehat g(j)\gets\frac{\phi_j(0,A)}{\pi(A)}\,g(x_A)$\;
\If{$A\ne B$}{
$\widehat g(j)\gets\widehat g(j)+\frac{\phi_j(A\wedge B,A\vee B)}{q(A\wedge B,A\vee B)}\bigl(g(x_{A\vee B})-g(x_{A\wedge B})\bigr)$\;
}
}
\Return{$\widehat g$}\;
}
\end{algorithm}

\paragraph{Second moment.}
The first lemma holds for the general estimator of the previous subsection.

\begin{lemma}[Second moment]
\label{lem:second-moment}
Under the assumptions of \Cref{lem:unbiased-estimator},
\[
\E\langle\pi,\widehat g^2\rangle
\le
2\sum_e\frac{H(e)^2}{q(e)}\sum_j\pi(j)\phi_j(e)^2,
\]
where the outer sum runs over edges with $q(e)>0$.
\end{lemma}

The right side pairs two factors for each edge: the load it carries relative to its revelation probability, which depends only on $\pi$, and its squared increment. The next lemma bounds the per-edge loads; it is a discrete-integral computation, and the factor $\Lambda$ in the interval bound is the price of the kernel normalization.

\begin{lemma}[Edge loads]
\label{lem:edge-loads}
For every rank $a$,
\[
\frac{\sum_{j\ge a}\pi(j)\phi_j(0,a)^2}{q(0,a)}
=
O\Bigl(\frac{\pi(a)}{\PiMass[1,a]}\Bigr),
\]
and for every pair of ranks $1\le a<b$,
\[
\frac{\sum_{j\ge b}\pi(j)\phi_j(a,b)^2}{q(a,b)}
=
O\Bigl(\Lambda
\frac{\pi(a)\pi(b)}{\PiMass(a,b]\PiMass[a,b]}\Bigr).
\]
\end{lemma}

The last lemma sums these weights against the squared increments. For monotone $g$, a layer-cake decomposition reduces the sum to edges crossing a single cut of the chain, whose total weight is a harmonic sum of order $\Lambda$; general $g$ follows from the Jordan decomposition, which is where the variation of $g$ enters.

\begin{lemma}[Ordered cut bound]
\label{lem:cut-bound}
Let $H(a,b)=g(x_b)-g(x_a)$ with $g(x_0)=0$, and let $V=|g(x_1)|+\TV(g)$ be the total variation of $g$ on the chain extended by the anchor. Then
\[
\sum_a H(0,a)^2\frac{\pi(a)}{\PiMass[1,a]}
+
\sum_{1\le a<b}H(a,b)^2
\frac{\pi(a)\pi(b)}{\PiMass(a,b]\PiMass[a,b]}
=
O(\Lambda V^2).
\]
\end{lemma}

Under the hypotheses of the theorem, $V\le2$, so combining the last three lemmas gives the second-moment bound of \Cref{thm:endpoint-flow}.

\paragraph{Self-bound.}
The self-bound is a direct computation: for every target $j\ge a$, $\pi(j)\phi_j(0,a)/q(0,a)\le1$, and for every target $j\ge b$, $\pi(j)\phi_j(a,b)/q(a,b)\le2\Lambda$. Thus the anchor and interval terms contribute at most $1$ and $2\Lambda$ times their respective increment magnitudes. For $g$ with values in $[0,1]$, every increment, including the anchor increments, has magnitude at most one, giving $1+2\Lambda\le3\Lambda$. The proofs of the lemmas of this section are in \Cref{app:endpoint-flow}.

\section{The separated monotone problem}
\label{sec:separated}

We now show how the endpoint flow estimator solves the separated monotone problem. Querying an action reveals $\ell_t^{\uparrow}$ and $\ell_t^{\downarrow}$ there separately, so the estimator can be applied to each component in its own orientation, and by the monotone-sign property both estimates are nonnegative. Let $R_T^{(2)}$ denote the minimax expected regret of two-point learners over oblivious loss sequences of the separated monotone problem and expert recommendations. This section proves that $R_T^{(2)}$ is $O(\log(|\cF|T)\sqrt{T\log|\cF|})$ and $\Omega(\sqrt T\log\min\{|\cF|,\sqrt T\})$.

\subsection{Algorithm and upper bound}

Let $u_t$ be the current expert mixture, and let $p_t(j)=\sum_{f:\,j_t(f)=j}u_t(f)$ be the induced distribution on active ranks, which aggregates experts recommending the same action. Smooth directly on active actions:
\[
\pi_t=\Bigl(1-\frac1T\Bigr)p_t+\frac1T\operatorname{Unif}(X_t).
\]
Run the endpoint flow sampler using $\pi_t$. Read the active chain forward for $\ell_t^{\uparrow}$ and backward for $\ell_t^{\downarrow}$; reading the chain backward replaces each rank $j$ by $m_t+1-j$ in $\pi_t$, in the pair, and in the output of the estimator. Reversing the order preserves the unordered interval and its $\pi_t$-mass, so the same paid pair can be inserted into both oriented estimators. Each estimate is nonnegative, and their sum is unbiased for the total loss. \Cref{alg:separated} summarizes the procedure.

\begin{algorithm}[t]
\caption{Endpoint flow with exponential weights for the separated monotone problem}
\label{alg:separated}
\KwIn{learning rate $\eta>0$}
Set $L_0(f)=0$ for every $f\in\cF$\;
\For{$t=1,\ldots,T$}{
Receive the recommendations and form the active chain $X_t$\;
Set $u_t(f)\propto\exp(-\eta L_{t-1}(f))$, $p_t(j)=\sum_{f:\,j_t(f)=j}u_t(f)$, and $\pi_t=(1-\frac1T)p_t+\frac1T\operatorname{Unif}(X_t)$\;
$(A_t,B_t)\gets{}$\Pair{$\pi_t$}\;
Play $x_{t,A_t}$ and $x_{t,B_t}$, and observe $\ell_t^{\uparrow}$ and $\ell_t^{\downarrow}$ at both\;
$\widehat\ell_t^{\uparrow}\gets{}$\Est{$\pi_t,A_t,B_t,\ell_t^{\uparrow}(x_{t,A_t}),\ell_t^{\uparrow}(x_{t,B_t})$} on the chain read forward\;
$\widehat\ell_t^{\downarrow}\gets{}$\Est{$\pi_t,A_t,B_t,\ell_t^{\downarrow}(x_{t,A_t}),\ell_t^{\downarrow}(x_{t,B_t})$} on the chain read backward\;
Set $L_t(f)=L_{t-1}(f)+\widehat\ell_t^{\uparrow}(j_t(f))+\widehat\ell_t^{\downarrow}(j_t(f))$ for every $f\in\cF$\;
}
\end{algorithm}

Because the estimates are nonnegative, the standard second-order exponential-weights bound applies.

\begin{lemma}[Exponential weights for nonnegative estimates]
\label{lem:nonnegative-ew}
Let $u_t$ be the exponential-weights distribution generated by nonnegative loss estimates $\widehat\ell_t(f)\ge0$:
\[
u_t(f)=\frac{\exp(-\eta L_{t-1}(f))}{\sum_{f'}\exp(-\eta L_{t-1}(f'))},
\qquad
L_t(f)=L_{t-1}(f)+\widehat\ell_t(f).
\]
Then for every $f^\star$,
\[
\sum_{t=1}^T\langle u_t,\widehat\ell_t\rangle
-
\sum_{t=1}^T\widehat\ell_t(f^\star)
\le
\frac{\log|\cF|}{\eta}
+
\frac\eta2\sum_{t=1}^T\langle u_t,\widehat\ell_t^2\rangle.
\]
\end{lemma}

\begin{theorem}[Separated two-point upper bound]
\label{thm:separated-upper}
Let $\Lambda=1+\log(|\cF|T)$. For every oblivious loss sequence of the separated monotone problem and arbitrary current recommendations, \Cref{alg:separated} with $\eta=\sqrt{\log|\cF|/(\Lambda^2T)}$ satisfies
\[
\E\Reg_T(\cF)
=
O\bigl(\log(|\cF|T)\sqrt{T\log|\cF|}\bigr).
\]
\end{theorem}

Each component takes values in $[0,1/2]$ and has total variation at most $1/2$, so the theorem applies to it in its own orientation, and the sum of the two reconstructions has expert-weighted second moment $O(\Lambda_t^2)$, where $\Lambda_t=1+\log(m_tT)\le1+\log(|\cF|T)$. Pathwise nonnegativity permits the second-order exponential-weights bound of the lemma above, and the stated learning rate makes both of its terms $O(\Lambda\sqrt{T\log|\cF|})$. The smoothing itself costs at most one in cumulative loss.

\subsection{Lower bound: a hidden dyadic path}

We now complement the upper bound with a lower bound that holds even for fixed recommendations.

\begin{theorem}[Two-point lower bound]
\label{thm:dyadic-lower}
For all $|\cF|,T\ge2$ and every two-point learner, some oblivious instance of the separated monotone problem forces
\[
\E\Reg_T(\cF)=\Omega\bigl(\sqrt T\log\min\{|\cF|,\sqrt T\}\bigr).
\]
In the instance every expert recommends one fixed action on every round. Every loss has total variation at most one, and observing only the losses at the two queried actions reveals the same information as observing both components, so the bound also holds for the BV problem.
\end{theorem}

The hard instance uses threshold losses $\ell_t(x)=\frac{1-x}{2}+\frac12\1\{Z_t<x\}$ on $[0,1]$, so a query at $x$ reveals only whether the random threshold $Z_t$ lies below $x$, and a simultaneous pair reveals only which of three bins contains $Z_t$. Experts sit at the centers of the $2^k$ dyadic cells of depth $k\approx\log_2\min\{|\cF|,\sqrt T\}$. The adversary hides a leaf $\theta$ and tilts the law of $Z_t$ by $\delta\asymp1/\sqrt T$ times the sum of the Haar functions along the dyadic path to $\theta$; this lowers the mean loss of the expert at $\theta$ by order $\delta k$ relative to any action chosen without knowledge of $\theta$.

Averaged over $\theta$, the tilt moves the probability of a bin of length $r$ by only $O(\delta\sqrt r)$, so each simultaneous pair carries $O(\delta^2)$ relative entropy about $\theta$. Over $T$ rounds the learner gathers $O(1)$ information, cannot locate $\theta$, and loses $\Omega(\delta k)$ per round, a total of $\Omega(\sqrt T\log\min\{|\cF|,\sqrt T\})$. Proofs are in \Cref{app:separated}.

\section{The BV problem}
\label{sec:bv}

We now turn to the BV problem, where querying $x$ reveals only the loss $\ell_t(x)$. Endpoint flow remains unbiased for the loss vector, but the estimate is signed, so the exponential-weights analysis of the separated monotone problem no longer applies.

\subsection{Algorithm and upper bound}

For signed estimates we use mirror descent with a regularizer that combines negative entropy with a log barrier. Its regret bound requires only a coordinatewise bound on $w_t(f)|\widehat\ell_t(f)|$, which is exactly the self-bound of endpoint flow in \Cref{thm:endpoint-flow}. Write $\Delta^\circ(\cF)$ for the set of distributions on $\cF$ with full support, and let
\[
D_\Psi(u,w)=\Psi(u)-\Psi(w)-\langle\nabla\Psi(w),u-w\rangle
\]
be the Bregman divergence generated by a differentiable convex regularizer $\Psi$.

\begin{proposition}[Robust mirror descent]
\label{prop:robust-md}
Let $\widehat\ell_1,\ldots,\widehat\ell_T$ be signed loss estimates on $\cF$, let $\mu>0$, and run the update
\[
w_{t+1}=\arg\min_{w\in\Delta^\circ(\cF)}
\bigl\{D_\Psi(w,w_t)+\eta\langle w,\widehat\ell_t\rangle\bigr\}
\]
with the regularizer
\[
\Psi(w)=\sum_f w(f)\log w(f)-\mu\sum_f\log w(f).
\]
Suppose that $\max_f w_t(f)|\widehat\ell_t(f)|\le\beta$ on every round. If $\mu\ge2\eta\beta$, then for every $u\in\Delta^\circ(\cF)$,
\[
\sum_{t=1}^T\langle w_t-u,\widehat\ell_t\rangle
\le
\frac{D_\Psi(u,w_1)}{\eta}
+\eta\sum_{t=1}^T\sum_f w_t(f)\widehat\ell_t(f)^2.
\]
\end{proposition}

\Cref{alg:bv} runs the endpoint flow sampler and feeds the signed loss estimate to this update.

\begin{algorithm}[t]
\caption{Endpoint flow with robust mirror descent for the BV problem}
\label{alg:bv}
\KwIn{learning rate $\eta>0$, barrier weight $\mu>0$}
Set $w_1(f)=1/|\cF|$ for every $f\in\cF$, and let $\Psi(w)=\sum_f w(f)\log w(f)-\mu\sum_f\log w(f)$\;
\For{$t=1,\ldots,T$}{
Receive the recommendations and form the active chain $X_t$\;
Set $p_t(j)=\sum_{f:\,j_t(f)=j}w_t(f)$ and $\pi_t=(1-\frac1T)p_t+\frac1T\operatorname{Unif}(X_t)$\;
$(A_t,B_t)\gets{}$\Pair{$\pi_t$}\;
Play $x_{t,A_t}$ and $x_{t,B_t}$, and observe $\ell_t$ at both\;
$\widehat\ell_t\gets{}$\Est{$\pi_t,A_t,B_t,\ell_t(x_{t,A_t}),\ell_t(x_{t,B_t})$}\;
Set $w_{t+1}=\arg\min_{w\in\Delta^\circ(\cF)}\bigl\{D_\Psi(w,w_t)+\eta\sum_f w(f)\widehat\ell_t(j_t(f))\bigr\}$\;
}
\end{algorithm}

\begin{theorem}[BV two-point upper bound]
\label{thm:BV-upper}
Let $\Lambda=1+\log(|\cF|T)$. For every oblivious loss sequence of the BV problem and arbitrary current recommendations, \Cref{alg:bv} with $\eta=\sqrt{\log|\cF|/(\Lambda^2T)}$ and $\mu=12\eta\Lambda$ satisfies
\[
\E\Reg_T(\cF)
=
O\bigl(\log(|\cF|T)\sqrt{T\log|\cF|}+|\cF|\log(|\cF|T)\log T\bigr)
=
\wtO\bigl(\sqrt T+|\cF|\bigr).
\]
\end{theorem}

The additive $|\cF|$ comes from the barrier diameter required to stabilize a signed importance-weighted estimator; it is not merely an artifact of the analysis, as the next lower bound shows.

\subsection{Lower bounds}

\begin{theorem}[Needle lower bound]
\label{thm:needle}
For every two-point algorithm for the BV problem, there is an oblivious loss sequence of the BV problem such that
\[
\E\Reg_T(\cF)
=
\Omega\bigl(\min\{|\cF|,T\}\bigr).
\]
\end{theorem}

The instance is literal search. Let $M=\min\{|\cF|,T\}$ and hide one action $i^\star$ with loss zero; every other candidate has loss $1/2$ on every round. Until $i^\star$ is queried, every feedback history is identical. Two simultaneous queries inspect at most two new candidates, so the expected discovery time is $\Omega(M)$. Every loss has total variation at most one.

The two-point lower bound of \Cref{thm:dyadic-lower} also applies to the BV problem, and together with the needle it determines the two-point BV rate up to logarithmic factors.

\begin{corollary}[Two-point BV rate]
\label{cor:BV-two-point-rate}
For all $|\cF|,T\ge2$ and every two-point algorithm for the BV problem, some oblivious loss sequence of the BV problem forces
\[
\E\Reg_T(\cF)=\Omega\Bigl(\sqrt T\log\min\{|\cF|,\sqrt T\}+\min\{|\cF|,T\}\Bigr).
\]
Together with the upper bound above, the two-point minimax regret for the BV problem is $\wtTheta(\sqrt T+|\cF|)$ when $|\cF|\le T$.
\end{corollary}

The remaining proofs are in \Cref{app:bv}.

\section{Dynamic pricing with expert advice}
\label{sec:menu-application}

We apply online learning with ordered experts to adversarial
dynamic pricing. We show that a menu with two purchase options
implements two paid price observations on the same buyer,
allowing the seller to compete with the best price-recommending
expert. The classical fixed-price benchmark follows as a
special case.

\subsection{Model and benchmarks}

On each round $t$, a seller faces a fresh risk-neutral buyer
with private valuation $v_t\in[0,1]$. An oblivious adversary
fixes the valuation sequence $v_1,\ldots,v_T$ in advance;
valuations may vary arbitrarily across rounds.

The seller has access to a finite class $\cF$ of experts.
Before the seller acts on round $t$, each expert $f\in\cF$
recommends a price $a_{t,f}\in[0,1]$. These recommendations
may vary arbitrarily across rounds. For example, an expert
may recommend a price based on publicly observed contextual
information. The recommendation sequences are also fixed
by the oblivious adversary in advance.

Posting price $p$ on round $t$ yields revenue
\[
r_t(p)=p\1\{v_t\ge p\}.
\]
Our benchmark is the best expert in hindsight, namely,
\[
\max_{f\in\cF}\sum_{t=1}^T r_t(a_{t,f}).
\]
The benchmark follows one expert throughout the horizon,
although that expert may recommend different prices on
different rounds.

Rather than posting a single price, the seller may offer
a finite menu
\[
\mathcal M_t\subseteq[0,1]\times[0,1],
\qquad (0,0)\in\mathcal M_t.
\]
Each entry $(\alpha,\tau)$ specifies an allocation probability
$\alpha$ and an expected payment $\tau$. The buyer selects
an entry maximizing $v_t\alpha-\tau$, breaking ties in favor
of the larger allocation probability. The seller observes
the selected entry $(\alpha_t,\tau_t)$ and receives expected
revenue $\tau_t$.

We measure expected revenue regret against the best expert:
\[
\Reg_T^{\cF}
=
\max_{f\in\cF}\sum_{t=1}^T r_t(a_{t,f})
-
\E\sum_{t=1}^T\tau_t.
\]
The learner may change its menu each round, while the
benchmark follows a single expert. 

\paragraph{Fixed-price pricing as a special case.}
When each expert recommends a constant price
$a_{t,f_p}=p$ on every round, the expert-advice benchmark
reduces to the best fixed price in the expert class.
Taking all prices in $[0,1]$ gives the classical
fixed-price benchmark \citep*{kleinberg2003value}:
\[
\max_{p\in[0,1]}\sum_{t=1}^T r_t(p).
\]

Importantly, allowing a fixed menu instead of a fixed
posted price does not improve this benchmark. For any
fixed finite menu $\mathcal M$, the allocation probability
selected by the buyer is nondecreasing in the valuation.
At each threshold $p_j$, an allocation increase of $w_j$
raises the payment by $w_jp_j$. Consequently,
\[
\tau_{\mathcal M}(v)
=
\sum_j w_jp_j\1\{v\ge p_j\},
\qquad
w_j\ge0,\quad \sum_jw_j\le1.
\]
Therefore,
\[
\sum_{t=1}^T\tau_{\mathcal M}(v_t)
=
\sum_jw_j\sum_{t=1}^T r_t(p_j)
\le
\max_{p\in[0,1]}\sum_{t=1}^T r_t(p).
\]
Since posted prices are themselves menus, the optimal
fixed menu and the optimal fixed posted price have the
same cumulative expected revenue.

This equivalence concerns the fixed-price benchmark;
it does not assert that an arbitrary expert class
dominates all fixed menus.

\subsection{Reduction to ordered experts via two-option menus}

\paragraph{Separated monotone losses.}
On round $t$, consider the set of distinct prices
recommended by the experts,
\[
\mathcal P_t=\{a_{t,f}:f\in\cF\},
\]
ordered by their numerical values. Define the normalized
loss
\[
\ell_t(p)=\frac{1-r_t(p)}{2}.
\]
It admits the decomposition
\[
\ell_t^\uparrow(p)
=
\frac p2\1\{v_t<p\},
\qquad
\ell_t^\downarrow(p)
=
\frac{1-p}{2},
\qquad
\ell_t=\ell_t^\uparrow+\ell_t^\downarrow.
\]
For every $v_t$, the first component is nondecreasing
in price and the second is nonincreasing. Both take
values in $[0,1/2]$, and
$\ell_t^\uparrow(0)=\ell_t^\downarrow(1)=0$.
Thus dynamic pricing with expert recommendations is an
instance of the separated monotone ordered-expert problem.

An acceptance decision at price $p$ determines both
loss components at that price. The remaining challenge
is to obtain acceptance feedback at two prices from
the same buyer, while preserving the learner's payoff.

\paragraph{Implementing two observations with a menu.}
Run \Cref{alg:separated} on the experts' current price
recommendations. Its endpoint-flow sampler draws two
prices $P_t,Q_t\in\mathcal P_t$, each with the intended
smoothed marginal distribution $\pi_t$.

Write
\[
p_t^-=\min\{P_t,Q_t\},
\qquad
p_t^+=\max\{P_t,Q_t\}.
\]
Offer the menu
\[
\mathcal M_t
=
\left\{
(0,0),
\left(\frac12,\frac{p_t^-}{2}\right),
\left(1,\frac{p_t^-+p_t^+}{2}\right)
\right\}.
\]
The first purchase option gives a half chance of
allocation at expected payment $p_t^-/2$.
Upgrading to the second gives an additional half
chance of allocation at additional expected payment
$p_t^+/2$.

Under the stated tie-breaking rule, the buyer selects
\[
(\alpha_t,\tau_t)=
\begin{cases}
(0,0), & v_t<p_t^-,\\[2pt]
(1/2,p_t^-/2), & p_t^-\le v_t<p_t^+,\\[2pt]
(1,(p_t^-+p_t^+)/2), & v_t\ge p_t^+.
\end{cases}
\]
The choice therefore reveals whether the buyer would
accept neither price, only the lower price, or both.
This recovers the two revenues $r_t(P_t),r_t(Q_t)$
and hence the separated loss observations required
by the algorithm.

Moreover, the menu's expected payment satisfies
\[
\tau_t=\frac{r_t(P_t)+r_t(Q_t)}2,
\]
so that
\[
\frac{\ell_t(P_t)+\ell_t(Q_t)}2
=
\frac{1-\tau_t}{2}.
\]
Thus the menu implements exactly the feedback and
payoff of the two-point protocol. Since both sampled
prices have marginal $\pi_t$, the menu's expected
revenue equals that of the intended price mixture.

In particular, revenue regret against the best expert
is twice the corresponding normalized loss regret.
The reduction applies even when expert recommendations
change and reorder arbitrarily across rounds.

\begin{theorem}[Pricing with menus]
\label{thm:pricing-expert-advice}
Online mechanisms offering two purchase options and an
outside option achieve the following expected
revenue-regret guarantees:
\begin{enumerate}[label=(\roman*),nosep,leftmargin=*]
\item \textbf{Expert advice.}
For any finite class $\cF$ of at least two experts
with oblivious price recommendations $a_{t,f}\in[0,1]$,
\[
\max_{f\in\cF}\sum_{t=1}^T r_t(a_{t,f})
-
\E\sum_{t=1}^T\tau_t
\le
O\!\left(
\log(|\cF|T)\sqrt{T\log|\cF|}
\right).
\]

\item \textbf{No expert advice.}
Against the best fixed posted price,
\[
\max_{p\in[0,1]}\sum_{t=1}^T r_t(p)
-
\E\sum_{t=1}^T\tau_t
\le
O\!\left(\sqrt T\,\log^{3/2}T\right).
\]
\end{enumerate}
Both mechanisms are incentive compatible and
individually rational in expected utility.
\end{theorem}

The expert-advice guarantee follows from
\Cref{alg:separated} and the separated monotone
regret bound. For the fixed-price case, construct
$T+1$ constant-price experts on the grid
$\{0,1/T,\ldots,1\}$. Rounding any price down to
this grid loses at most $1/T$ revenue per round,
and hence at most one over the horizon.
The full proof appears in \Cref{sec:menu-proof}.

\section{One-point and full-information feedback}
\label{sec:one-point-full-info}

To calibrate the two-point results, we compare them with two other protocols. Under \emph{one-point} feedback, the learner chooses one active action $A_t$, incurs $\ell_t(A_t)$, and receives the problem-specific observation at that point; regret is defined as for two points, with the paid average replaced by $\ell_t(A_t)$. Under \emph{full information}, the learner chooses its action before seeing the current loss and then observes the full loss vector on the active recommendations.

\subsection{One-point feedback in the separated monotone problem}

\begin{theorem}[Separated one-point lower bound]
\label{thm:separated-one-lower}
For every randomized one-point learner and all sufficiently large $T$, there is a deterministic oblivious instance of the separated monotone problem with constant recommendations such that
\[
\E\Reg_T(\cF)
=
\Omega\bigl(\min\{\sqrt{|\cF|T},T^{2/3}\}\bigr).
\]
\end{theorem}

The instance hides one slightly better action inside a monotone component. Keeping the component monotone requires a compensating slope, which limits the number of hidden candidates to $T^{1/3}$.

\subsection{One-point feedback in the BV problem}

\begin{theorem}[One-point BV feedback]
\label{thm:BV-onepoint}
For $2\le|\cF|\le T$, the minimax one-point regret in the BV problem satisfies $R_T^{(1)}=\Theta(\sqrt{|\cF|T})$, and the lower bound holds with losses that also belong to the separated monotone problem.
\end{theorem}

The upper bound is the standard adversarial bandit guarantee with $|\cF|$ arms. The lower bound hides one slightly better action among $|\cF|$ candidates; all actions share a common random baseline on each round, so every loss has small total variation.

\subsection{Full information: order alone does not help}

\begin{theorem}[Full-information benchmark]
\label{thm:full-info}
For $\log_2|\cF|\le T$, the minimax full-information regret in both problems satisfies $R_T^{\mathrm{full}}=\Theta(\sqrt{T\log|\cF|})$.
The lower bound holds even when every expert recommends one fixed action on every round, and its argument does not use what the learner observes.
\end{theorem}

Exponential weights gives the upper bound, and a nested-block construction gives the lower bound. Proofs are in \Cref{app:one-point}.

\clearpage
\bibliographystyle{plainnat}
\bibliography{references}

@inproceedings{alon2015,
  title={Online learning with feedback graphs: Beyond bandits},
  author={Alon, Noga and Cesa-Bianchi, Nicol{\`o} and Dekel, Ofer and Koren, Tomer},
  booktitle={Proceedings of the 28th Conference on Learning Theory (COLT)},
  year={2015}
}

@inproceedings{audibert2009,
  title={Minimax policies for adversarial and stochastic bandits},
  author={Audibert, Jean-Yves and Bubeck, S{\'e}bastien},
  booktitle={Proceedings of the 22nd Conference on Learning Theory (COLT)},
  year={2009}
}

@article{bubeck2012,
  title={Regret analysis of stochastic and nonstochastic multi-armed bandit problems},
  author={Bubeck, S{\'e}bastien and Cesa-Bianchi, Nicol{\`o}},
  journal={Foundations and Trends in Machine Learning},
  volume={5},
  number={1},
  pages={1--122},
  year={2012}
}

@article{auer2002nonstochastic,
  title={The nonstochastic multiarmed bandit problem},
  author={Auer, Peter and Cesa-Bianchi, Nicol{\`o} and Freund, Yoav and Schapire, Robert E.},
  journal={SIAM Journal on Computing},
  volume={32},
  number={1},
  pages={48--77},
  year={2002}
}

@inproceedings{mcmahan2009tighter,
  title={Tighter bounds for multi-armed bandits with expert advice},
  author={McMahan, H. Brendan and Streeter, Matthew},
  booktitle={Proceedings of the 22nd Conference on Learning Theory (COLT)},
  year={2009}
}

@book{cesa2006prediction,
  title={Prediction, Learning, and Games},
  author={Cesa-Bianchi, Nicol{\`o} and Lugosi, G{\'a}bor},
  publisher={Cambridge University Press},
  year={2006}
}

@book{lattimore2020bandit,
  title={Bandit Algorithms},
  author={Lattimore, Tor and Szepesv{\'a}ri, Csaba},
  publisher={Cambridge University Press},
  year={2020}
}

@article{freund1997decision,
  title={A decision-theoretic generalization of on-line learning and an application to boosting},
  author={Freund, Yoav and Schapire, Robert E.},
  journal={Journal of Computer and System Sciences},
  volume={55},
  number={1},
  pages={119--139},
  year={1997}
}

@article{littlestone1994weighted,
  title={The weighted majority algorithm},
  author={Littlestone, Nick and Warmuth, Manfred K.},
  journal={Information and Computation},
  volume={108},
  number={2},
  pages={212--261},
  year={1994}
}

@inproceedings{flaxman2005online,
  title={Online convex optimization in the bandit setting: Gradient descent without a gradient},
  author={Flaxman, Abraham D. and Kalai, Adam Tauman and McMahan, H. Brendan},
  booktitle={Proceedings of the 16th Annual ACM-SIAM Symposium on Discrete Algorithms (SODA)},
  year={2005}
}

@inproceedings{agarwal2010optimal,
  title={Optimal algorithms for online convex optimization with multi-point bandit feedback},
  author={Agarwal, Alekh and Dekel, Ofer and Xiao, Lin},
  booktitle={Proceedings of the 23rd Conference on Learning Theory (COLT)},
  year={2010}
}

@article{duchi2015optimal,
  title={Optimal rates for zero-order convex optimization: The power of two function evaluations},
  author={Duchi, John C. and Jordan, Michael I. and Wainwright, Martin J. and Wibisono, Andre},
  journal={IEEE Transactions on Information Theory},
  volume={61},
  number={5},
  pages={2788--2806},
  year={2015}
}

@article{shamir2017optimal,
  title={An optimal algorithm for bandit and zero-order convex optimization with two-point feedback},
  author={Shamir, Ohad},
  journal={Journal of Machine Learning Research},
  volume={18},
  number={52},
  pages={1--11},
  year={2017}
}

@inproceedings{mannor2011bandits,
  title={From bandits to experts: On the value of side-observations},
  author={Mannor, Shie and Shamir, Ohad},
  booktitle={Advances in Neural Information Processing Systems 24 (NeurIPS)},
  year={2011}
}

@inproceedings{seldin2014prediction,
  title={Prediction with limited advice and multiarmed bandits with paid observations},
  author={Seldin, Yevgeny and Bartlett, Peter L. and Crammer, Koby and Abbasi-Yadkori, Yasin},
  booktitle={Proceedings of the 31st International Conference on Machine Learning (ICML)},
  year={2014}
}

@article{yue2012k,
  title={The {$K$}-armed dueling bandits problem},
  author={Yue, Yisong and Broder, Josef and Kleinberg, Robert and Joachims, Thorsten},
  journal={Journal of Computer and System Sciences},
  volume={78},
  number={5},
  pages={1538--1556},
  year={2012}
}

@inproceedings{kleinberg2003value,
  title={The value of knowing a demand curve: Bounds on regret for online posted-price auctions},
  author={Kleinberg, Robert and Leighton, Tom},
  booktitle={Proceedings of the 44th Annual IEEE Symposium on Foundations of Computer Science (FOCS)},
  year={2003}
}

@inproceedings{kleinberg2008multi,
  title={Multi-armed bandits in metric spaces},
  author={Kleinberg, Robert and Slivkins, Aleksandrs and Upfal, Eli},
  booktitle={Proceedings of the 40th Annual ACM Symposium on Theory of Computing (STOC)},
  year={2008}
}

@inproceedings{yu2011unimodal,
  title={Unimodal bandits},
  author={Yu, Jia Yuan and Mannor, Shie},
  booktitle={Proceedings of the 28th International Conference on Machine Learning (ICML)},
  year={2011}
}

@inproceedings{combes2014unimodal,
  title={Unimodal bandits: Regret lower bounds and optimal algorithms},
  author={Combes, Richard and Prouti{\`e}re, Alexandre},
  booktitle={Proceedings of the 31st International Conference on Machine Learning (ICML)},
  year={2014}
}

@article{huh2009nonparametric,
  title={A nonparametric asymptotic analysis of inventory planning with censored demand},
  author={Huh, Woonghee Tim and Rusmevichientong, Paat},
  journal={Mathematics of Operations Research},
  volume={34},
  number={1},
  pages={103--123},
  year={2009}
}

@inproceedings{foster2016learning,
  title={Learning in games: Robustness of fast convergence},
  author={Foster, Dylan J. and Li, Zhiyuan and Lykouris, Thodoris and Sridharan, Karthik and Tardos, {\'E}va},
  booktitle={Advances in Neural Information Processing Systems 29 (NeurIPS)},
  year={2016}
}

@inproceedings{agarwal2017corralling,
  title={Corralling a band of bandit algorithms},
  author={Agarwal, Alekh and Luo, Haipeng and Neyshabur, Behnam and Schapire, Robert E.},
  booktitle={Proceedings of the 30th Conference on Learning Theory (COLT)},
  year={2017}
}

@inproceedings{wei2018more,
  title={More adaptive algorithms for adversarial bandits},
  author={Wei, Chen-Yu and Luo, Haipeng},
  booktitle={Proceedings of the 31st Conference on Learning Theory (COLT)},
  year={2018}
}

@inproceedings{bubeck2018sparsity,
  title={Sparsity, variance and curvature in multi-armed bandits},
  author={Bubeck, S{\'e}bastien and Cohen, Michael B. and Li, Yuanzhi},
  booktitle={Proceedings of Algorithmic Learning Theory (ALT)},
  year={2018}
}

@article{hastings1970monte,
  title={Monte {C}arlo sampling methods using {M}arkov chains and their applications},
  author={Hastings, W. Keith},
  journal={Biometrika},
  volume={57},
  number={1},
  pages={97--109},
  year={1970}
}

@inproceedings{liu2021optimal,
  author    = {Liu, Allen and Paes Leme, Renato and Schneider, Jon},
  title     = {Optimal Contextual Pricing and Extensions},
  booktitle = {Proceedings of the 2021 ACM-SIAM Symposium
               on Discrete Algorithms (SODA)},
  year      = {2021},
  pages     = {1059--1078},
  doi       = {10.1137/1.9781611976465.66}
}

@inproceedings{xu2024pricing,
  author    = {Xu, Jianyu and Wang, Yu-Xiang},
  title     = {Pricing with Contextual Elasticity and
               Heteroscedastic Valuation},
  booktitle = {Proceedings of the 41st International Conference
               on Machine Learning},
  series    = {Proceedings of Machine Learning Research},
  volume    = {235},
  pages     = {55286--55304},
  year      = {2024},
  publisher = {PMLR},
  url       = {https://proceedings.mlr.press/v235/xu24x.html}
}

@article{balcan2024menus,
  author  = {Balcan, Maria-Florina and Beyhaghi, Hedyeh},
  title   = {New Guarantees for Learning Revenue Maximizing Menus
             of Lotteries and Two-Part Tariffs},
  journal = {Transactions on Machine Learning Research},
  year    = {2024},
  url     = {https://openreview.net/forum?id=mhawjZcmrJ}
}

@misc{zuo2026menus,
  author        = {Zuo, Shiliang},
  title         = {The Power of Menus in Dynamic Pricing:
                   Optimal Regret and Equivalence Results},
  year          = {2026},
  eprint        = {2604.24093},
  archivePrefix = {arXiv},
  primaryClass  = {cs.GT},
  note          = {arXiv:2604.24093v3},
  url           = {https://arxiv.org/abs/2604.24093v3}
}

\newpage
\appendix

\section{Proofs for endpoint flow}
\label{app:endpoint-flow}

This appendix proves the lemmas used in the proof of \Cref{thm:endpoint-flow}.

\subsection{Unbiasedness and second moment}

Write $\widehat g(j)=z_1(j)+z_2(j)$, where
\[
z_1(j)=\sum_{a}\1\{E_1=(0,a)\}\frac{\phi_j(0,a)}{q(0,a)}H(0,a),
\qquad
z_2(j)=\sum_{1\le a<b}\1\{E_2=(a,b)\}\frac{\phi_j(a,b)}{q(a,b)}H(a,b)
\]
are the two sums in the definition of $\widehat g$. At most one indicator in each sum equals one.

\begin{proof}[Proof of \Cref{lem:unbiased-estimator}]
Since $E_1=(0,a)$ with probability $q(0,a)$,
\[
\E z_1(j)=\sum_{a:\,q(0,a)>0}q(0,a)\frac{\phi_j(0,a)}{q(0,a)}H(0,a)=\sum_a\phi_j(0,a)H(0,a),
\]
where the last step uses that $q(0,a)>0$ whenever $\phi_j(0,a)>0$. In the same way, $\E z_2(j)=\sum_{1\le a<b}\phi_j(a,b)H(a,b)$. Adding the two and using the flow identity $g(x_j)=\sum_e\phi_j(e)H(e)$ gives $\E\widehat g(j)=g(x_j)$. If $g$ is nonnegative and nondecreasing, then every increment is nonnegative, including the anchor increments $H(0,a)=g(x_a)$; since the flows are nonnegative, $z_1(j)$ and $z_2(j)$ are nonnegative for every realization.
\end{proof}

\begin{proof}[Proof of \Cref{lem:second-moment}]
By $(z_1+z_2)^2\le2z_1^2+2z_2^2$,
\[
\E\widehat g(j)^2
\le
2\E z_1(j)^2+2\E z_2(j)^2
=
2\sum_a\frac{\phi_j(0,a)^2}{q(0,a)}H(0,a)^2
+
2\sum_{1\le a<b}\frac{\phi_j(a,b)^2}{q(a,b)}H(a,b)^2,
\]
with both sums over edges of positive revelation probability. Multiplying by $\pi(j)$, summing over $j$, and exchanging the target and edge sums proves the lemma. No independence between $E_1$ and $E_2$ is needed.
\end{proof}

\subsection{Kernel validity}

\begin{proof}[Proof of \Cref{lem:kernel-validity}]
Symmetry of $\Gamma$ gives detailed balance. For a fixed $b$, the off-diagonal row mass is
\[
\frac1{4\Lambda}
\left(
\sum_{a<b}\frac{\pi(a)}{\PiMass[a,b]}
+
\sum_{a>b}\frac{\pi(a)}{\PiMass[b,a]}
\right).
\]
Both sums are harmonic. For the first, set $S_a=\PiMass[a,b]$. Since
\[
\frac{\pi(a)}{S_a}
=
\frac{S_a-S_{a+1}}{S_a}
\le
\log S_a-\log S_{a+1},
\]
we get
\[
\sum_{a<b}\frac{\pi(a)}{\PiMass[a,b]}
\le
\log\frac{\PiMass[1,b]}{\pi(b)}
\le
\log\frac{m}{\xi}
\le \Lambda.
\]
The second sum is identical by symmetry. Thus the off-diagonal row mass is at most $1/2$.
\end{proof}

\subsection{Random-path edge probabilities}

Fix target $j$. Starting from the anchor $0$, at node $a<j$ the path jumps to $b\in\{a+1,\ldots,j\}$ with probability $\pi(b)/\PiMass(a,j]$.

\begin{lemma}[Edge-use probabilities]
\label{lem:edge-use}
The probabilities that the path uses the anchor edge $(0,a)$ and the interval edge $(a,b)$ are
\[
\phi_j(0,a)=\frac{\pi(a)}{\PiMass[1,j]}
\quad(a\le j),
\qquad
\phi_j(a,b)=
\frac{\pi(a)\pi(b)}{\PiMass[a,j]\PiMass(a,j]}
\quad(1\le a<b\le j).
\]
The edge-use probability is zero outside $0\le a<b\le j$.
\end{lemma}

\begin{proof}
The path reaches $a\le j$ with probability $\pi(a)/\PiMass[a,j]$. This is proved inductively using
\[
\frac{\pi(r)}{\PiMass[r,j]\PiMass(r,j]}
=
\frac1{\PiMass(r,j]}-\frac1{\PiMass[r,j]}.
\]
Multiplying the reach probability of $a$ by the next-step probability $\pi(b)/\PiMass(a,j]$ gives the formula. The anchor case is the same calculation starting at $0$.
\end{proof}

\subsection{Discrete integral and edge-load estimates}

We use the following elementary estimate.

\begin{lemma}[Discrete integral bounds]
\label{lem:discrete-integrals}
Let $0\le s_{b-1}<s_b<s_{b+1}<\cdots$ and $\delta_j=s_j-s_{j-1}$. Then
\[
\sum_{j\ge b}\frac{\delta_j}{s_j^2}\le \frac{2}{s_b}.
\]
For every $\alpha>0$,
\[
\sum_{j\ge b}\frac{\delta_j}{s_j^2(\alpha+s_j)^2}
=
O\Bigl(\frac{1}{s_b(\alpha+s_b)^2}\Bigr).
\]
\end{lemma}

\begin{proof}
For the first claim, $\delta_b\le s_b$ and the function $s\mapsto s^{-2}$ is decreasing, so
\[
\sum_{j\ge b}\frac{\delta_j}{s_j^2}
\le\frac1{s_b}+\int_{s_b}^{\infty}\frac{ds}{s^2}
=\frac2{s_b}.
\]
For the second claim, apply the same comparison to $s\mapsto s^{-2}(\alpha+s)^{-2}$ and use $(\alpha+s)^{-2}\le(\alpha+s_b)^{-2}$ for $s\ge s_b$. This gives an upper bound of $2/[s_b(\alpha+s_b)^2]$.
\end{proof}

\begin{proof}[Proof of \Cref{lem:edge-loads}]
For the anchor edge $(0,a)$, substitute $\phi_j(0,a)=\pi(a)/\PiMass[1,j]$ and $q(0,a)=\pi(a)$, and apply the first part of \Cref{lem:discrete-integrals}, starting at index $a$, to the cumulative masses $s_j=\PiMass[1,j]$. For an interval edge, set $s_j=\PiMass(a,j]$ and $\alpha=\pi(a)$, apply the second discrete-integral bound, then divide by $q(a,b)=2\Gamma(a,b)=\pi(a)\pi(b)/(2\Lambda\PiMass[a,b])$.
\end{proof}

\subsection{Ordered cut bound}

\begin{proof}[Proof of \Cref{lem:cut-bound}]
First suppose $g$ is nondecreasing on the chain including the anchor, with $g(x_0)=0$ and $g(x_m)=V$. Since $0\le H(a,b)\le V$,
\[
H(a,b)^2\le VH(a,b)=V\int_0^V\1\{g(x_a)\le z<g(x_b)\}\,dz.
\]
For fixed $z\in[0,V)$, monotonicity makes the lower-level set an initial segment $\{0,\ldots,k\}$ with $k<m$. Thus the contributing edges cross a single ordered cut. For the anchor edges,
\[
\sum_{b>k}\frac{\pi(b)}{\PiMass[1,b]}
\le\sum_{b=1}^m\frac{\pi(b)}{\PiMass[1,b]}
\le1+\log\frac1{\pi(1)}\le\Lambda.
\]
For an interval edge, the identity
\[
\frac{\pi(a)}{\PiMass(a,b]\PiMass[a,b]}
=\frac1{\PiMass(a,b]}-\frac1{\PiMass[a,b]}
\]
gives, for $k\ge1$ and $b>k$,
\[
\sum_{a=1}^k\frac{\pi(a)\pi(b)}{\PiMass(a,b]\PiMass[a,b]}
=\pi(b)\left(\frac1{\PiMass(k,b]}-\frac1{\PiMass[1,b]}\right)
\le\frac{\pi(b)}{\PiMass(k,b]}.
\]
Summing over $b>k$ gives at most $1+\log(1/\pi(k+1))\le\Lambda$ by the same harmonic bound. When $k=0$ there are no interval edges crossing the cut. The total cut weight is therefore at most $2\Lambda$, and integrating over an interval of length $V$ yields $O(\Lambda V^2)$.

For a general $g$, take the Jordan decomposition $g=g^+-g^-$ with $g^\pm(x_0)=0$ and both nondecreasing. Then
\[
H^2\le2(H^+)^2+2(H^-)^2,
\]
and $V^++V^-=|g(x_1)|+\TV(g)=V$, the total variation of $g$ on the chain extended by the anchor. Applying the monotone bound to both pieces proves the lemma.
\end{proof}

\subsection{Coordinatewise self-bound}

\begin{lemma}[Self-bound]
\label{lem:endpoint-selfbound-proof}
For every realization and every active rank $j$,
\[
\pi(j)|\widehat g(j)|
\le
3\Lambda\max_{0\le a<b\le m}|g(x_b)-g(x_a)|,
\]
where $g(x_0)=0$. In particular, the right side is at most $3\Lambda$ when $g$ takes values in $[0,1]$.
\end{lemma}

\begin{proof}
For the anchor term, if $A=a\le j$, then
\[
z_1(j)=\frac{g(x_a)-g(x_0)}{\PiMass[1,j]},
\]
so multiplication by $\pi(j)$ gives at most the increment magnitude; if $A>j$, then $z_1(j)=0$. For the interval term, when $a<b\le j$,
\[
\frac{\phi_j(a,b)}{2\Gamma(a,b)}
=
\frac{2\Lambda\PiMass[a,b]}{\PiMass[a,j]\PiMass(a,j]}.
\]
Because $\PiMass[a,b]\le\PiMass[a,j]$ and $\pi(j)\le\PiMass(a,j]$, multiplying by $\pi(j)$ leaves at most $2\Lambda$ times the observed increment. Adding the two terms and using $\Lambda\ge1$ proves the claim.
\end{proof}

\section{Proofs for the separated monotone problem}
\label{app:separated}

This appendix contains the proofs for \Cref{sec:separated}.

\subsection{Exponential weights for nonnegative estimates}

\begin{proof}[Proof of \Cref{lem:nonnegative-ew}]
Let
\[
\Phi(L)=-\frac1\eta\log\left(\frac1{|\cF|}\sum_f e^{-\eta L(f)}\right).
\]
For the exponential-weights distribution $u_t=\nabla\Phi(L_{t-1})$ and a nonnegative estimate $\widehat\ell_t$, the inequality $e^{-x}\le1-x+x^2/2$ for $x\ge0$ gives
\[
\langle u_t,\widehat\ell_t\rangle
\le
\Phi(L_t)-\Phi(L_{t-1})
+\frac\eta2\langle u_t,\widehat\ell_t^2\rangle.
\]
Summing over $t$ and using
\[
\Phi(L_T)\le L_T(f^\star)+\frac{\log|\cF|}{\eta}
\]
proves the lemma.
\end{proof}

\subsection{Separated two-point upper bound}

\begin{proof}[Proof of \Cref{thm:separated-upper}]
Let $u_t$ be the expert mixture before round $t$, and let $p_t$ be its induced distribution on the distinct active actions. Smooth with $\xi=1/T$,
\[
\pi_t=(1-\xi)p_t+\xi\operatorname{Unif}(X_t),
\]
and write
\[
\Lambda_t=1+\log\frac{m_t}{\xi}\le1+\log(|\cF|T).
\]
Run the endpoint flow sampler with target marginal $\pi_t$.

Read $\ell_t^{\uparrow}$ in the forward order from the anchor below the chain and $\ell_t^{\downarrow}$ in the reversed order from an anchor above the chain. Reversal preserves every unordered interval and its $\pi_t$-mass, so the conductances and hence the sampled pair are identical in the two orientations. The same paid pair therefore feeds both estimators.

Let $\widehat\ell_t^{\uparrow}$ and $\widehat\ell_t^{\downarrow}$ denote the two reconstructed vectors on active actions, and set
\[
\widehat\ell_t=\widehat\ell_t^{\uparrow}+\widehat\ell_t^{\downarrow}.
\]
Each component takes values in $[0,1/2]$ and has total variation at most $1/2$, and read in its own orientation it is nondecreasing. By \Cref{thm:endpoint-flow}, both reconstructed vectors are pathwise nonnegative and unbiased. Therefore $\widehat\ell_t\ge0$ coordinatewise and
\[
\E_t[\widehat\ell_t(j)]=\ell_t(x_{t,j}).
\]
Moreover, by $(x+y)^2\le2x^2+2y^2$ and the second-moment bound of \Cref{thm:endpoint-flow},
\[
\E_t\sum_j\pi_t(j)\widehat\ell_t(j)^2
=O(\Lambda_t^2).
\]
Since $p_t(j)\le\pi_t(j)/(1-\xi)\le2\pi_t(j)$ for $\xi\le1/2$, aggregation from actions back to experts gives
\[
\E_t\sum_f u_t(f)\widehat\ell_t(f)^2
=O(\Lambda_t^2).
\]

Apply \Cref{lem:nonnegative-ew}, take conditional expectations, and use unbiasedness. Because both paid endpoints have marginal $\pi_t$, their expected paid average equals $\langle\pi_t,\ell_t\rangle$. Replacing $p_t$ by $\pi_t$ costs at most $\xi$ per round since losses lie in $[0,1]$. Hence, for every comparator expert $f^\star$,
\[
\E\Reg_T(\cF)
\le
\frac{\log|\cF|}{\eta}
+O\Bigl(\eta\sum_{t=1}^T\Lambda_t^2\Bigr)
+\xi T.
\]
The smoothing cost is $\xi T=1$. Since $\Lambda_t\le\Lambda$, the middle term is $O(\eta\Lambda^2T)$, and the choice $\eta=\sqrt{\log|\cF|/(\Lambda^2T)}$ makes both remaining terms equal to $O(\Lambda\sqrt{T\log|\cF|})$. This yields
\[
\E\Reg_T(\cF)
=
O\bigl(\log(|\cF|T)\sqrt{T\log|\cF|}\bigr).
\]
\end{proof}

\subsection{Two-point lower bound: a hidden dyadic path}
\label{app:dyadic-lower}

This subsection proves \Cref{thm:dyadic-lower}. Throughout, the action space is $[0,1]$, so $\bot=0$ and $\top=1$.

\paragraph{Threshold losses.} For $z\in[0,1]$ let
\[
\ell_z(x)=\frac{1-x}{2}+\frac12\1\{z<x\},
\qquad
\ell^{\downarrow}(x)=\frac{1-x}{2},
\qquad
\ell_z^{\uparrow}(x)=\frac12\1\{z<x\}.
\]
The component $\ell^{\downarrow}$ is nonincreasing with values in $[0,1/2]$ and $\ell^{\downarrow}(1)=0$; the component $\ell_z^{\uparrow}$ is nondecreasing with values in $\{0,1/2\}$ and $\ell_z^{\uparrow}(0)=0$. Hence $\ell_z$ is a loss of the separated monotone problem, and its total variation on any chain is at most one, so it is also a BV loss. Because $\ell^{\downarrow}$ does not depend on $z$, observing both components at $x$, or only $\ell_z(x)$, is equivalent to observing the bit $\1\{z<x\}$. A pair $0\le x\le y\le1$ therefore reveals exactly which of the bins $[0,x)$, $[x,y)$, $[y,1]$ contains $z$, the middle bin being empty when $x=y$.

\paragraph{Small depth.} Let $M=\min\{|\cF|,\lfloor\sqrt T\rfloor\}$. If $M<2^{11}$, then $\min\{|\cF|,\sqrt T\}<2^{11}$, so $\log\min\{|\cF|,\sqrt T\}<8$, and it suffices to force regret $\Omega(\sqrt T)$. Place one expert at $0$, one at $1$, and the remaining experts anywhere. On each round, independently, draw a Rademacher sign $\sigma_t$ and use $\ell_t(x)=x/2$ if $\sigma_t=+1$ and $\ell_t(x)=(1-x)/2$ if $\sigma_t=-1$. These are losses of the separated monotone problem with total variation $1/2$. Every action has conditional expected loss $1/4$ given the past, so every learner, whatever it observes, has expected loss $T/4$, while the better of the experts at $0$ and $1$ has loss $T/4-\frac14\bigl|\sum_t\sigma_t\bigr|$. Since $\E|\sum_t\sigma_t|\ge c\sqrt T$, the regret is $\Omega(\sqrt T)$.

\paragraph{Parameters.} Assume from now on that $M\ge2^{11}$, and set
\[
k=\lfloor\log_2M\rfloor\ge11,
\qquad
K=2^k\le M,
\qquad
\delta=\frac{c_0}{\sqrt T},
\qquad
c_0=\frac{1}{128\sqrt2}.
\]
Since $\lfloor\sqrt T\rfloor\ge\sqrt T/2$, we have $k\ge\log_2\min\{|\cF|,\sqrt T\}-2\ge\frac{11}{13}\log_2\min\{|\cF|,\sqrt T\}$. Place one expert at each point of
\[
\mathcal C=\Bigl\{\frac{i+1/2}{K}:\ i=0,1,\ldots,K-1\Bigr\},
\]
the centers of the dyadic cells of depth $k$, and place any remaining experts at other points of $[0,1]$. The argument below allows the learner to query arbitrary points of $[0,1]$.

\paragraph{Haar functions and tents.} For a dyadic interval $I=[a,a+w]$ with left half $I_{\mathrm L}$ and right half $I_{\mathrm R}$, let
\[
\psi_I=\frac{\1_{I_{\mathrm L}}-\1_{I_{\mathrm R}}}{w},
\qquad
\tau_I(x)=\int_0^x\psi_I(z)\,dz=\1\{x\in I\}\,\frac{\min\{x-a,\,a+w-x\}}{w}.
\]
Thus $\tau_I$ is a tent of height $1/2$ supported on $I$, it is $1/w$-Lipschitz, and it vanishes at $0$ and $1$. For $\theta\in\mathcal C$ and $0\le d<k$, let $I_d(\theta)$ be the dyadic interval of width $2^{-d}$ containing $\theta$, and set
\[
G_\theta=\sum_{d=0}^{k-1}\psi_{I_d(\theta)},
\qquad
H_\theta=\sum_{d=0}^{k-1}\tau_{I_d(\theta)},
\]
so that $H_\theta(x)=\int_0^xG_\theta(z)\,dz$.

\paragraph{The random instance.} Draw $\Theta$ uniformly from $\mathcal C$. Given $\Theta=\theta$, draw $Z_1,\ldots,Z_T$ independently from the density $\rho_\theta=1-\delta G_\theta$, and use the losses $\ell_t=\ell_{Z_t}$. Each Haar function integrates to zero, and $|G_\theta|\le\sum_{d<k}2^d<K$ with $\delta K\le c_0<1/4$, so $\rho_\theta$ is a density with values in $[3/4,5/4]$. Its distribution function is $x-\delta H_\theta(x)$, so the mean loss is
\[
\E_\theta\,\ell_{Z_t}(x)=\frac{1-x}{2}+\frac12\bigl(x-\delta H_\theta(x)\bigr)=\frac12-\frac\delta2H_\theta(x).
\]
The null environment draws $Z_1,\ldots,Z_T$ independently and uniformly from $[0,1]$.

\begin{lemma}[Tents along a random path]
\label{lem:tents}
Let $\Theta$ be uniform on $\mathcal C$. Then
\begin{enumerate}
\item $0\le H_\theta(x)\le k/2$ for every $\theta$ and $x$;
\item $\E H_\Theta(x)<1$ for every fixed $x$;
\item $\E H_\Theta(\Theta)=k/4$;
\item $\E\bigl(H_\Theta(y)-H_\Theta(x)\bigr)^2\le64(y-x)$ for $0\le x<y\le1$, and in particular $\E H_\Theta(x)^2\le64\min\{x,1-x\}$.
\end{enumerate}
\end{lemma}

\begin{proof}
The first claim holds because $H_\theta$ is a sum of $k$ tents with values in $[0,1/2]$.

For the second claim, fix $x$ and a depth $d$, and let $w=2^{-d}$. Only the depth-$d$ dyadic interval containing $x$ can contribute a nonzero tent at $x$, and it is the ancestor $I_d(\Theta)$ with probability $w$, because it contains a fraction $w$ of the leaf centers. Its contribution is at most $1/2$, so the expected contribution of depth $d$ is at most $w/2$, and summing over the $k$ depths gives $\E H_\Theta(x)\le\sum_{d=0}^{k-1}2^{-d-1}=1-2^{-k}<1$.

For the third claim, condition on the depth-$d$ ancestor $I=[a,a+w]$ of $\Theta$. Then $\Theta$ is uniform over the $2^{k-d}\ge2$ leaf centers in $I$, half of them in each half of $I$. The leaf centers in the left half are symmetric about the quarter point $a+w/4$, and on the left half $\tau_I(u)=(u-a)/w$, so a center and its reflection about $a+w/4$ have tent heights summing to $1/2$; a center at the quarter point itself has height $1/4$. The same holds in the right half. Hence the conditional mean of $\tau_I(\Theta)$ is $1/4$, and summing over the $k$ depths gives $k/4$.

For the last claim, let $r=y-x$ and $X_d=\tau_{I_d(\Theta)}(y)-\tau_{I_d(\Theta)}(x)$, and let $w=2^{-d}$. The difference $X_d$ is nonzero only if $I_d(\Theta)$ contains $x$ or $y$, which has probability at most $2w$, and $|X_d|\le\min\{r/w,1/2\}$ because the tent is $1/w$-Lipschitz with values in $[0,1/2]$. Hence $\|X_d\|_2\le\sqrt{2w}\,\min\{r/w,1/2\}$, where $\|\cdot\|_2$ is the $L^2$ norm over $\Theta$. By Minkowski's inequality, splitting the dyadic widths at $r$,
\[
\bigl\|H_\Theta(y)-H_\Theta(x)\bigr\|_2
\le
\sqrt2\,r\sum_{w\ge r}w^{-1/2}
+\sum_{w<r}\sqrt{w/2}
\le
5\sqrt r+3\sqrt r,
\]
since both are geometric sums with ratio $\sqrt2$ whose largest terms are at most $\sqrt2\,r\cdot r^{-1/2}$ and $\sqrt{r/2}$, respectively. Squaring gives the bound $64r$. Since every tent vanishes at $0$ and $1$, applying it to the intervals $[0,x]$ and $[x,1]$ gives the final inequality.
\end{proof}

\begin{lemma}[Three-bin information]
\label{lem:three-bin}
For a pair $0\le x\le y\le1$, let $Q_0$ and $Q_\theta$ be the laws of the bin of $Z_t$ among $[0,x)$, $[x,y)$, $[y,1]$ under the null environment and under the hidden leaf $\theta$. Then
\[
\frac1K\sum_{\theta\in\mathcal C}\KL(Q_0\|Q_\theta)\le256\,\delta^2.
\]
\end{lemma}

\begin{proof}
The bin probabilities are $(x,\,y-x,\,1-y)$ under the null and
\[
\bigl(x-\delta H_\theta(x),\ y-x+\delta H_\theta(x)-\delta H_\theta(y),\ 1-y+\delta H_\theta(y)\bigr)
\]
under $\theta$. Since $\rho_\theta\ge3/4$, every bin probability under $\theta$ is at least $3/4$ of its null probability. Bounding the relative entropy by the $\chi^2$ divergence,
\[
\KL(Q_0\|Q_\theta)
\le
\frac{4\delta^2}{3}
\left(
\frac{H_\theta(x)^2}{x}
+\frac{\bigl(H_\theta(y)-H_\theta(x)\bigr)^2}{y-x}
+\frac{H_\theta(y)^2}{1-y}
\right),
\]
where a term with a vanishing denominator is omitted, since its numerator vanishes as well. By \Cref{lem:tents}, the average over $\theta$ of each of the three quotients is at most $64$, which gives $\frac43\cdot3\cdot64\,\delta^2=256\,\delta^2$.
\end{proof}

\begin{proof}[Proof of \Cref{thm:dyadic-lower}]
Fix a two-point learner. For the analysis only, after the learner chooses its pair $(A_t,B_t)$, let $Y_t$ equal $A_t$ or $B_t$ with probability $1/2$ each, independently of everything else. Then the conditional expectation of $\ell_t(Y_t)$ given the pair and $Z_t$ is the paid average $\frac12(\ell_t(A_t)+\ell_t(B_t))$. Let $P_0$ and $P_\theta$ be the laws of the transcript, consisting of the learner's internal randomness, its pairs, the observed bins, and $Y_1,\ldots,Y_T$, under the null environment and under the hidden leaf $\theta$.

\emph{Information.} On every round the learner chooses its pair from the past transcript in the same way under every environment, and the auxiliary choice of $Y_t$ does not depend on the environment. By the chain rule for relative entropy and \Cref{lem:three-bin},
\[
\frac1K\sum_{\theta}\KL(P_0\|P_\theta)\le256\,T\delta^2=256\,c_0^2=\frac1{128}.
\]
By Pinsker's inequality and Jensen's inequality,
\[
\frac1K\sum_{\theta}\TVdist(P_0,P_\theta)
\le
\sqrt{\frac1{2}\cdot\frac1{128}}
=\frac1{16}.
\]

\emph{Tent height collected by the learner.} For each $\theta$ let $S_\theta=\sum_{t=1}^TH_\theta(Y_t)$, a function of the transcript with $0\le S_\theta\le Tk/2$ by \Cref{lem:tents}. Under $P_0$ the transcript does not depend on $\theta$, so averaging over $\theta$ and using the second claim of \Cref{lem:tents} gives $\frac1K\sum_\theta\E_0S_\theta\le T$. Changing measure,
\[
\frac1K\sum_\theta\E_\theta S_\theta
\le
\frac1K\sum_\theta\Bigl(\E_0S_\theta+\frac{Tk}{2}\,\TVdist(P_0,P_\theta)\Bigr)
\le
T+\frac{Tk}{32}.
\]

\emph{Regret.} The best expert in hindsight has loss at most that of the expert at $\theta$. Since $Y_t$ is chosen before $Z_t$ is drawn, the mean-loss formula gives
\[
\E_\theta\Reg_T
\ge
\E_\theta\sum_{t=1}^T\bigl(\ell_t(Y_t)-\ell_t(\theta)\bigr)
=
\frac\delta2\bigl(T\,H_\theta(\theta)-\E_\theta S_\theta\bigr).
\]
Averaging over $\theta$ and using the third claim of \Cref{lem:tents},
\[
\frac1K\sum_\theta\E_\theta\Reg_T
\ge
\frac{\delta T}{2}\Bigl(\frac k4-1-\frac k{32}\Bigr)
\ge
\frac{\delta Tk}{16},
\]
where the last step uses $k\ge11$. Since $\delta=c_0/\sqrt T$ and $k\ge\frac{11}{13}\log_2\min\{|\cF|,\sqrt T\}\ge\log\min\{|\cF|,\sqrt T\}$, the average regret is $\Omega(\sqrt T\log\min\{|\cF|,\sqrt T\})$. All randomness of the instance, including $\Theta$, is drawn before play, so some realization is an oblivious instance with at least this regret.
\end{proof}

\section{Proofs for the BV problem}
\label{app:bv}

This appendix contains the proofs for \Cref{sec:bv}.

\subsection{Robust mirror descent for signed estimates}

\begin{proof}[Proof of \Cref{prop:robust-md}]
The barrier term makes $\Psi$ a Legendre function on $\Delta^\circ(\cF)$, so the minimizer $w_{t+1}$ exists, has full support, and satisfies the first-order condition
\[
\nabla\Psi(w_{t+1})=\nabla\Psi(w_t)-\eta\widehat\ell_t-\lambda_t\mathbf 1
\]
for some scalar $\lambda_t$. Since $\langle u-w_{t+1},\mathbf 1\rangle=0$, the three-point identity for Bregman divergences gives
\[
\eta\langle w_{t+1}-u,\widehat\ell_t\rangle
=
D_\Psi(u,w_t)-D_\Psi(u,w_{t+1})-D_\Psi(w_{t+1},w_t).
\]
Adding $\eta\langle w_t-w_{t+1},\widehat\ell_t\rangle$ to both sides,
\[
\eta\langle w_t-u,\widehat\ell_t\rangle
\le
D_\Psi(u,w_t)-D_\Psi(u,w_{t+1})+S_t,
\]
where
\[
S_t
=
\sup_{v\in\mathbb R_{>0}^{\cF}}
\{\eta\langle w_t-v,\widehat\ell_t\rangle-D_\Psi(v,w_t)\}
\]
and we have relaxed the simplex constraint. Because $\Psi$ is separable, writing $y_f=v(f)/w_t(f)$ gives
\[
D_\Psi(v,w_t)
=
\sum_f w_t(f)\,\varphi(y_f)
+
\mu\sum_f(y_f-1-\log y_f),
\qquad
\varphi(y)=y\log y-y+1,
\]
so $S_t\le\sum_f\sup_{y>0}F_f(y)$ with
\[
F_f(y)
=
w_t(f)\bigl(c_f(y-1)-\varphi(y)\bigr)-\mu(y-1-\log y),
\qquad
c_f=-\eta\widehat\ell_t(f).
\]
We show $\sup_{y>0}F_f(y)\le w_t(f)c_f^2$.

\emph{Case $0<y\le2$.} Since $\varphi(1)=\varphi'(1)=0$ and $\varphi''(s)=1/s\ge1/2$ for $s\le2$, Taylor's theorem gives $\varphi(y)\ge(y-1)^2/4$. Dropping the nonnegative barrier term,
\[
F_f(y)\le w_t(f)\Bigl(c_f(y-1)-\frac{(y-1)^2}{4}\Bigr)\le w_t(f)c_f^2.
\]

\emph{Case $y>2$.} Here
\[
F_f'(y)=w_t(f)c_f-w_t(f)\log y-\mu\Bigl(1-\frac1y\Bigr)
<
w_t(f)|c_f|-\frac\mu2
\le0,
\]
because $\log y>0$, $1-1/y\ge1/2$, and $\mu\ge2\eta\beta\ge2w_t(f)|c_f|$. Hence $F_f$ is decreasing on $[2,\infty)$ and $\sup_{y>2}F_f(y)\le F_f(2)\le w_t(f)c_f^2$ by the first case.

Therefore
\[
S_t\le\eta^2\sum_f w_t(f)\widehat\ell_t(f)^2.
\]
Summing the three-point inequality over $t$, discarding $-D_\Psi(u,w_{T+1})\le0$, and dividing by $\eta$ proves the proposition.
\end{proof}

\subsection{BV two-point upper bound}

\begin{proof}[Proof of \Cref{thm:BV-upper}]
Run endpoint flow over the loss vector $\ell_t$, anchored below the chain. Let $w_t$ be the expert mixture produced by the robust mirror-descent update, aggregate it on active actions as
\[
p_t(j)=\sum_{f:\,j_t(f)=j}w_t(f),
\]
and smooth
\[
\pi_t=(1-\xi)p_t+\xi\operatorname{Unif}(X_t).
\]
Here $\xi=1/T$. Let $\Lambda_t=1+\log(m_t/\xi)$; since $m_t\le|\cF|$, we have $\Lambda_t\le\Lambda$. Write $\widehat\ell_t(f)=\widehat\ell_t(j_t(f))$ for the estimate at the recommendation of $f$. Since $\ell_t$ takes values in $[0,1]$ and $\TV(\ell_t)\le1$, \Cref{thm:endpoint-flow} gives
\[
\E_t\sum_j\pi_t(j)\widehat\ell_t(j)^2
=
O(\Lambda^2).
\]
Since $p_t(j)\le\pi_t(j)/(1-\xi)\le2\pi_t(j)$,
\[
\E_t\sum_fw_t(f)\widehat\ell_t(f)^2
=
O(\Lambda^2).
\]
Likewise the self-bound and $w_t(f)\le2\pi_t(j_t(f))$ give
\[
\max_f w_t(f)|\widehat\ell_t(f)|\le6\Lambda.
\]
Thus \Cref{prop:robust-md} applies with $\beta=6\Lambda$ and $\mu=12\eta\Lambda$.

Use the interior comparator
\[
u=(1-\eps)e_{f^\star}+\eps\mathbf1/|\cF|.
\]
Comparator smoothing costs at most $\eps T$ and action smoothing costs at most $\xi T$. For the regularizer diameter, the entropy part of $D_\Psi(u,w_1)$ is $\KL(u\|w_1)\le\log|\cF|$, and since $\sum_f u(f)/w_1(f)=|\cF|$, the barrier part equals
\[
\mu\sum_f\Bigl(\frac{u(f)}{w_1(f)}-1-\log\frac{u(f)}{w_1(f)}\Bigr)
=
-\mu\sum_f\log\bigl(|\cF|u(f)\bigr)
\le
\mu|\cF|\log(1/\eps),
\]
because $|\cF|u(f)=\eps$ for $f\ne f^\star$ and $|\cF|u(f^\star)\ge1$. Hence
\[
D_\Psi(u,w_1)
\le
\log|\cF|+12\eta\Lambda|\cF|\log(1/\eps).
\]
Taking expectations and using unbiasedness gives
\begin{align*}
\E\Reg_T(\cF)
&\le
\eps T+\xi T
+12\Lambda|\cF|\log(1/\eps)
+\frac{\log|\cF|}{\eta}
+O(\eta\Lambda^2T).
\end{align*}
Setting $\eps=\xi=1/T$ and $\eta=\sqrt{\log|\cF|/(\Lambda^2T)}$ gives the claimed bound.
\end{proof}

\subsection{Needle lower bound}

\begin{proof}[Proof of \Cref{thm:needle}]
Let $M=\min\{|\cF|,T\}$. Place one expert on each action $1\prec\cdots\prec M$ and duplicate arbitrary experts if $|\cF|>M$. Draw $i^\star$ uniformly from $[M]$ and set, for every round,
\[
\ell_t(i)=\frac12\1\{i\ne i^\star\}.
\]
This is realizable in the separated monotone problem: take
\[
\ell_t^{\uparrow}(i)=\frac12\1\{i>i^\star\},
\qquad
\ell_t^{\downarrow}(i)=\frac12\1\{i<i^\star\}.
\]
Both components satisfy the endpoint normalizations and take values in $[0,1/2]$.

Let $\tau$ be the first round on which $i^\star$ is among the two queried actions, with $\tau=T+1$ if this never occurs. Fix the learner's internal randomness and run it counterfactually on the all-$1/2$ feedback history. The first $s$ query pairs touch a deterministic set $S_s$ of at most $2s$ candidate actions. The real interaction agrees with this counterfactual trajectory until $i^\star$ is queried, so
\[
\Prb(\tau\le s)=\Prb(i^\star\in S_s)\le\frac{2s}{M}.
\]
Therefore
\[
\E\tau
\ge
\sum_{s=0}^{\lceil M/2\rceil-1}\left(1-\frac{2s}{M}\right)
\ge
\frac M4.
\]
Before discovery the learner cannot play the zero-loss action. Hence its expected cumulative loss, equal to its regret because the best expert incurs zero, is at least
\[
\frac12(\E\tau-1)
\ge
\frac M8-\frac12.
\]
Moreover, on the first round each played action has expected loss $(1-1/M)/2\ge1/4$, so the regret is also at least $1/4$. Since $\max\{1/4,M/8-1/2\}\ge M/24$ for every $M\ge2$, the regret is at least $M/24$.\claudecomment{New: the old proof stopped at $M/8-1/2$, which is negative for $M<4$, so it did not give $\Omega(\min\{|\cF|,T\})$; the first-round bound fixes this and simplifies the corollary proof below.}
Averaging over $i^\star$ leaves at least one fixed needle location with the same lower bound for the randomized learner.
\end{proof}

\subsection{The two-point BV rate}

\begin{proof}[Proof of \Cref{cor:BV-two-point-rate}]
Let $M=\min\{|\cF|,T\}$. \Cref{thm:dyadic-lower} gives a BV instance with $\E\Reg_T\ge\rho:=c_1\sqrt T\log\min\{|\cF|,\sqrt T\}$ for a universal constant $c_1>0$, and the proof of \Cref{thm:needle} gives one with $\E\Reg_T\ge M/24$. The worse of the two instances forces at least $\max\{\rho,M/24\}\ge\rho/2+M/48$, which proves the corollary.
\end{proof}

\section{Proofs for one-point and full-information feedback}
\label{app:one-point}

This appendix contains the proofs for \Cref{sec:one-point-full-info}. We begin with the full-information benchmark.

\subsection{Full-information benchmark}

\begin{proof}[Proof of \Cref{thm:full-info}]
The upper bound is the standard exponential-weights guarantee $O(\sqrt{T\log|\cF|})$ for losses in $[0,1]$. For the lower bound, by Yao's principle it suffices to construct a random oblivious instance on which every learner has large expected regret; averaging then yields a fixed instance for every randomized learner. Let
\[
k=\lfloor\log_2|\cF|\rfloor,
\]
so that $1\le k\le T$ by the assumption $\log_2|\cF|\le T$.
Use the actions $1<\cdots<2^k$, so that $\bot=1$ and $\top=2^k$, place one expert permanently on each action, and place any remaining experts on action $1$; extra experts can only lower the comparator's loss. Partition the first $nk$ rounds into $k$ consecutive temporal blocks $\mathcal T_1,\ldots,\mathcal T_k$ of equal length
\[
n=\left\lfloor\frac{T}{k}\right\rfloor\ge1.
\]
Unused rounds, if any, have zero loss. The adversary samples all randomness below before the interaction.

Initially let $I_0=[2^k]$. In block $d$, the interval $I_{d-1}$ has two equal consecutive children $I_{d-1}^{\mathrm L}$ and $I_{d-1}^{\mathrm R}$ separated by a threshold $q_d$, with $1\le q_d<2^k$. On every round $t\in\mathcal T_d$, draw an independent Rademacher sign $\sigma_t$. If $\sigma_t=+1$, use the increasing step
\[
\ell_t(i)=\frac12\1\{i>q_d\};
\]
if $\sigma_t=-1$, use the decreasing step
\[
\ell_t(i)=\frac12\1\{i\le q_d\}.
\]
The increasing step is a component $\ell_t^{\uparrow}$ with $\ell_t^{\uparrow}(\bot)=0$, the decreasing step is a component $\ell_t^{\downarrow}$ with $\ell_t^{\downarrow}(\top)=0$, and each step has total variation $1/2$. Hence the instance belongs to both problems.

Let $\mathcal G_{t-1}$ be generated by the learner's internal randomness and $\sigma_1,\ldots,\sigma_{t-1}$. Everything the learner has observed before round $t$, under any feedback model, is $\mathcal G_{t-1}$-measurable, so its queries on round $t$ are $\mathcal G_{t-1}$-measurable, while $\sigma_t$ is independent of $\mathcal G_{t-1}$. Conditional on $\mathcal G_{t-1}$, every action has loss $1/2$ with probability $1/2$ and $0$ otherwise. Consequently the conditional expected paid average of a two-point learner is $1/4$ on every round, and so is the conditional expected loss of a one-point or full-information learner.

Let
\[
S_d=\sum_{t\in\mathcal T_d}\sigma_t.
\]
After block $d$, define $I_d$ to be whichever child of $I_{d-1}$ has smaller cumulative loss in that block, breaking ties arbitrarily. Every expert in $I_d$ has block-$d$ loss
\[
\frac n4-\frac{|S_d|}{4}.
\]
The intervals are nested,
\[
I_0\supset I_1\supset\cdots\supset I_k,
\]
and $I_k$ contains a single expert $i^\star$. Since this same expert belongs to the winning child at every level,
\[
\sum_{t=1}^{nk}\ell_t(i^\star)
=
\frac{nk}{4}-\frac14\sum_{d=1}^k|S_d|.
\]
The learner, meanwhile, has expected loss $nk/4$. Therefore
\[
\E\Reg_T
\ge
\frac14\sum_{d=1}^k\E|S_d|.
\]
A length-$n$ Rademacher sum satisfies $\E|S_d|\ge c\sqrt n$ for a universal constant $c>0$. Hence
\[
\E\Reg_T
\ge
c' k\sqrt n
\ge
c''\sqrt{Tk},
\]
using $n=\lfloor T/k\rfloor\ge T/(2k)$. Since $k\ge\frac12\log_2|\cF|$, this is $\Omega(\sqrt{T\log|\cF|})$.
\end{proof}

The two one-point constructions share a scalar common-noise perturbation and a hidden-coordinate testing lemma, which we record first. Both proofs use Yao's principle: it suffices to lower bound the average regret of deterministic learners under a random oblivious instance, and averaging then yields a fixed oblivious instance for every randomized learner.

\subsection{Smooth common-noise perturbation}

\label{app:perturbation}

\begin{lemma}[Smooth perturbation]
\label{lem:smooth-perturbation}
For every $\eps\in(0,1/4]$ there exist coupled random variables $Z,Z_\eps\in[1/4,1/2]$ with laws $P,Q_\eps$ such that $0\le Z-Z_\eps\le \eps/16$ almost surely,
\[
\E Z-\E Z_\eps=\eps/24,
\qquad
\KL(P\|Q_\eps)\le3\eps^2.
\]
\end{lemma}

\begin{proof}
Let $U\sim\operatorname{Unif}[0,1]$ and $\tau_\eps(u)=u-\eps u(1-u)$. Since $\tau_\eps'(u)=1-\eps+2\eps u\in[1-\eps,1+\eps]$, $\tau_\eps$ is an increasing bijection. Define
\[
Z=\frac14+\frac U4,
\qquad
Z_\eps=\frac14+\frac{\tau_\eps(U)}4.
\]
The pointwise bound follows from $u(1-u)\le1/4$, and
\[
\E(Z-Z_\eps)=\frac\eps4\E[U(1-U)]=\frac\eps{24}.
\]
After removing the common affine transformation, compare the uniform density on $[0,1]$ with the density $q_\eps(y)=1/\tau_\eps'(\tau_\eps^{-1}(y))$. For $\eps\le1/4$,
\[
\frac1{1+\eps}\le q_\eps(y)\le\frac1{1-\eps},
\qquad
|q_\eps(y)-1|\le\frac\eps{1-\eps}.
\]
Therefore $\chi^2(P\|Q_\eps)\le3\eps^2$, and $\KL(P\|Q_\eps)\le\chi^2(P\|Q_\eps)$.
\end{proof}

\subsection{One hidden favorable coordinate}
\label{app:testing}

\begin{lemma}[Testing lemma]
\label{lem:hidden-coordinate}
Fix $K\ge2$, $T\ge K$, and $\eps\in(0,1/4]$, and use the laws $P,Q_\eps$ from \Cref{lem:smooth-perturbation}. On each round, independently of the past, every action under the null has observation law $P$ and the same mean loss. Under alternative $\theta\in[K]$, only action $\theta$ changes to observation law $Q_\eps$ and has mean loss lower by $\Delta=\eps/24$. Then every one-point learner satisfies
\[
\frac1K\sum_{\theta=1}^K\E_\theta\Reg_T
\ge
\Delta T
\left(
1-\frac1K-\eps\sqrt{\frac{3T}{2K}}
\right).
\]
In particular, for sufficiently small universal $a>0$, $\eps=a\sqrt{K/T}$ gives $\Omega(\sqrt{KT})$ average regret.
\end{lemma}

\begin{proof}
Let
\[
N_\theta=\sum_{t=1}^T\1\{A_t=\theta\},
\qquad
n_\theta=\E_0N_\theta.
\]
The adaptive chain rule gives
\[
\KL(P_0\|P_\theta)
=n_\theta\KL(P\|Q_\eps)
\le3\eps^2n_\theta,
\]
because the observation law differs only when the learner selects $\theta$. Pinsker and $0\le N_\theta\le T$ imply
\[
\E_\theta N_\theta
\le
n_\theta+T\eps\sqrt{\frac{3n_\theta}{2}}.
\]
Since $\sum_\theta n_\theta=T$, Jensen yields
\[
\frac1K\sum_\theta\E_\theta N_\theta
\le
\frac TK+\eps T\sqrt{\frac{3T}{2K}}.
\]
Since the best action in hindsight has cumulative loss at most that of action $\theta$, independence of the current losses from the learner's past gives
\[
\E_\theta\Reg_T\ge\Delta\bigl(T-\E_\theta N_\theta\bigr).
\]
Averaging proves the lemma.
\end{proof}

\subsection{Separated one-point lower bound}

\begin{proof}[Proof of \Cref{thm:separated-one-lower}]
By Yao's principle it suffices to lower bound deterministic learners under a random oblivious instance. Let
\[
K=\min\{|\cF|,\lfloor T^{1/3}\rfloor\},
\qquad
\eps=a\sqrt{K/T},
\qquad
s=\eps/16,
\]
where $a$ is the constant from \Cref{lem:hidden-coordinate}. Use actions $\bot\prec1\prec\cdots\prec K$ and one constant expert at each positive action; place any remaining experts on action $1$.

Reveal before play the fixed nonincreasing component
\[
\ell^{\downarrow}(\bot)=sK,
\qquad
\ell^{\downarrow}(i)=s(K-i)\quad(i\in[K]).
\]
Since $K\le T^{1/3}$,
\[
sK=\frac{aK^{3/2}}{16\sqrt T}\le \frac a{16},
\]
so $\ell^{\downarrow}$ takes values in $[0,1/2]$ for small $a$, and $\ell^{\downarrow}(K)=0$.

Under the null, draw $Z_t$ independently and set
\[
\ell_t^{\uparrow,0}(\bot)=0,
\qquad
\ell_t^{\uparrow,0}(i)=Z_t-s(K-i).
\]
Under alternative $\theta$, use the coupled pair $(Z_t,Z_{t,\eps})$ and set
\[
\ell_t^{\uparrow,\theta}(\bot)=0,
\qquad
\ell_t^{\uparrow,\theta}(i)=
\begin{cases}
Z_{t,\eps}-s(K-\theta),&i=\theta,\\
Z_t-s(K-i),&i\ne\theta.
\end{cases}
\]
The pointwise bound $0\le Z_t-Z_{t,\eps}\le\eps/16$ and $s=\eps/16$ ensure that the increments into and out of $\theta$ remain nonnegative; all other positive-action increments equal $s$. Thus $\ell_t^{\uparrow,\theta}$ is nonnegative and nondecreasing, with values in $[0,1/2]$ after choosing $a$ small enough.

For $i\ne\theta$ the total loss is
\[
\ell^{\downarrow}(i)+\ell_t^{\uparrow,\theta}(i)=Z_t,
\]
while at $\theta$ it is $Z_{t,\eps}$. After adding the known downward component to the observed upward component, the observation law is $P$ off the favorable action and $Q_\eps$ at the favorable action, with gap $\Delta=\eps/24$. This invertible transformation preserves relative entropy. Even revealing $\ell^{\downarrow}$ before play only strengthens the learner. \Cref{lem:hidden-coordinate} therefore gives $\Omega(\sqrt{KT})$ regret.

If $|\cF|\le T^{1/3}$ then $K=|\cF|$ and this is $\Omega(\sqrt{|\cF|T})$; otherwise $K=\Theta(T^{1/3})$ and it is $\Omega(T^{2/3})$. Averaging over the random choice of $\theta$ and common-noise sequence yields a deterministic oblivious sequence for every randomized learner.
\end{proof}

\subsection{BV one-point lower bound}

We use \Cref{lem:smooth-perturbation,lem:hidden-coordinate} with $K=|\cF|\le T$ and $\eps=a\sqrt{K/T}$.

\begin{proof}[Proof of \Cref{thm:BV-onepoint}]
The standard adversarial $K$-armed bandit upper bound gives the matching upper rate. It remains to prove the lower bound.

Use fixed actions $[K]$ and one expert per action. Under the null, draw $Z_t$ independently and set
\[
\ell_t^0(i)=Z_t
\]
for every action $i$.
Under alternative $\theta$, use the coupled pair $(Z_t,Z_{t,\eps})$ and set
\[
\ell_t^\theta(i)=
\begin{cases}
Z_{t,\eps},&i=\theta,\\
Z_t,&i\ne\theta.
\end{cases}
\]
Every coordinate lies in $[1/4,1/2]$. The realized vector is constant except for one downward dip, so its total variation satisfies
\[
\TV(\ell_t^\theta)
\le
2|Z_t-Z_{t,\eps}|
\le
\frac\eps8
\le1.
\]
Thus the hard family lies in the BV model with variation at most one on every round.

We also verify explicit realizability in the separated monotone problem. In the null environment, take
\[
\ell_t^{\uparrow,0}(i)=Z_t\1\{i=K\},
\qquad
\ell_t^{\downarrow,0}(i)=Z_t\1\{i<K\}.
\]
For an alternative $\theta<K$, take
\[
\ell_t^{\uparrow,\theta}(i)=Z_t\1\{i>\theta\},
\]
\[
\ell_t^{\downarrow,\theta}(i)
=
Z_t\1\{i<\theta\}+Z_{t,\eps}\1\{i=\theta\}.
\]
For $\theta=K$, take
\[
\ell_t^{\uparrow,K}(i)=Z_{t,\eps}\1\{i=K\},
\qquad
\ell_t^{\downarrow,K}(i)=Z_t\1\{i<K\}.
\]
In every case the first component is nondecreasing and zero at the least action, the second is nonincreasing and zero at the greatest action, both are in $[0,1/2]$, and their sum is exactly $\ell_t^\theta$.

A one-point learner sees law $P$ at every ordinary action and law $Q_\eps$ at the hidden favorable action, with mean loss gap $\Delta=\eps/24$. \Cref{lem:hidden-coordinate} gives
\[
\frac1K\sum_\theta\E_\theta\Reg_T
=\Omega\bigl(\sqrt{|\cF|T}\bigr).
\]
Yao's principle and averaging yield a deterministic oblivious loss sequence for every randomized learner.
\end{proof}

\section{Proof of Dynamic Pricing Using Menus}
\label{sec:menu-proof}

\begin{proof}[Proof of \Cref{thm:pricing-expert-advice}]
\emph{(i) No expert advice.} Use $\mathcal P=\{j/T:j=0,\ldots,T\}$ and run \Cref{alg:separated} on its $T+1$ constant-price experts. These experts are constructed by the learner and require no external advice. The reduction above and \Cref{thm:separated-upper} give revenue regret against the best grid price at most
\[
O\bigl(\log(|\mathcal P|T)\sqrt{T\log|\mathcal P|}\bigr)
=O\bigl(\sqrt T\,\log^{3/2}T\bigr).
\]
For any price $p$, round down to $p'=\lfloor Tp\rfloor/T$. Whenever a buyer accepts $p$, it also accepts $p'$, and $r_t(p')\ge r_t(p)-1/T$ on every round. Thus the best grid price loses at most one in total revenue relative to the best price in $[0,1]$.

\emph{(ii) Expert advice.} Run the same algorithm on the current recommendations $a_{t,f}$. The menu choice reveals both sampled prices' acceptance decisions and supplies the separated component observations. Its expected payment is the two revenues' average, so revenue regret against the best expert is twice the algorithm's loss regret. \Cref{thm:separated-upper} gives the stated bound, even when the recommendations change and reorder arbitrarily.

In both cases the outside option and utility-maximizing menu choice give incentive compatibility and individual rationality in expected utility.
\end{proof}

\end{document}